\documentclass[lettersize,journal]{IEEEtran}
\usepackage{amsmath,amsfonts}
\usepackage{algorithmic}
\usepackage{algorithm}
\usepackage{array}
\usepackage[caption=false,font=normalsize,labelfont=sf,textfont=sf]{subfig}
\usepackage{textcomp}
\usepackage{stfloats}
\usepackage{url}
\usepackage{verbatim}
\usepackage{graphicx}
\usepackage{cite}
\usepackage{utfsym}
\usepackage{multirow}
\usepackage{hyperref}

\begin{document}

\title{CFNet: Optimizing Remote Sensing Change Detection through Content-Aware Enhancement}

\author{{Fan~Wu, Sijun~Dong, Xiaoliang~Meng*}
\thanks{This work was funded by the Major Program(JD) of Hubei Province (2023BAA025) and was also supported by the National Natural Science Foundation of China (NSFC) under Grant 41971352. (Corresponding author: Xiaoliang Meng.)}

\thanks{Fan Wu and Sijun Dong are with the School of Remote Sensing and Information Engineering, Wuhan University, Wuhan 430079, China (e-mail: wifibk@whu.edu.cn; dyzy41@whu.edu.cn).}

\thanks{Xiaoliang Meng is with the School of Remote Sensing and Information Engineering and Hubei Luojia Laboratory, Wuhan University, Wuhan 430079, China (e-mail: xmeng@whu.edu.cn).}
}

\markboth{IEEE Journal of Selected Topics in Applied Earth Observations and Remote Sensing}
{Wu\MakeLowercase{\textit{et al.}}: CFNet:A New Strategy For Remote Sensing Change Detection Based With Content-Aware}


\maketitle

\begin{abstract}
Change detection is a crucial and widely applied task in remote sensing, aimed at identifying and analyzing changes occurring in the same geographical area over time. Due to variability in acquisition conditions, bi-temporal remote sensing images often exhibit significant differences in image style. Even with the powerful generalization capabilities of DNNs, these unpredictable style variations between bi-temporal images inevitably affect the model’s ability to accurately detect changed areas. To address issue above, we propose the Content Focuser Network (CFNet), which takes content-aware strategy as a key insight. CFNet employs EfficientNet-B5 as the backbone for feature extraction. To enhance the model’s focus on the content features of images while mitigating the misleading effects of style features, we develop a constraint strategy that prioritizes the content features of bi-temporal images, termed Content-Aware. Furthermore, to enable the model to flexibly focus on changed and unchanged areas according to the requirements of different stages, we design a reweighting module based on the cosine distance between bi-temporal image features, termed Focuser. CFNet achieve outstanding performance across three well-known change detection datasets: CLCD (F1: 81.41\%, IoU: 68.65\%), LEVIR-CD (F1: 92.18\%, IoU: 85.49\%), and SYSU-CD (F1: 82.89\%, IoU: 70.78\%). The code and pretrained models of CFNet are publicly released at https://github.com/wifiBlack/CFNet.
\end{abstract}


\section{Introduction}
\IEEEPARstart{R}{emote} Sensing Change Detection (RSCD) is the process of identifying and quantifying changes in an object, phenomenon, or landscape by comparing images acquired through remote sensing technology at different times. This technique is crucial for understanding changes in land cover, urban growth, environmental shifts, and natural disasters\cite{asokan2019change}. The process involves comparing bi-temporal or multi-temporal satellite imagery or aerial photos, often with the help of advanced algorithms, to detect changes such as deforestation, urban expansion, or shoreline shifts\cite{liu2016deep}\cite{kleynhans2015detecting}. Additionally, RSCD encompasses various tasks, including bi-temporal binary change detection, bi-temporal multi-class change detection, and multi-temporal change analysis. In this study, we focus specifically on bi-temporal binary change detection, where the goal is to distinguish changed and unchanged areas between two input images.

Advancements in remote sensing technology have led to a rapid increase in the diversity of platforms, ranging from spaceborne (e.g., satellites, spacecraft) to airborne (e.g., drones, balloons) and ground-based platforms (e.g., sensing towers, vehicles) \cite{toth2016remote}\cite{zhang2023review}. This diversity has significantly increased the complexity of image style features. Even within the same platform, variations in atmospheric conditions, lighting, sensor calibration, and platform trajectory can introduce discrepancies between bi-temporal images\cite{zhao2022intelligent}\cite{ zhang2021curriculum}. 

While differences in style features are often emphasized in heterogeneous images (e.g., optical vs. SAR), similar variations in style can also occur within homogeneous multispectral images. These discrepancies in style can complicate the change detection process even when the images are from the same platform. Such style variations can exacerbate the challenge of isolating meaningful content changes, as they introduce unnecessary interference that confounds the model’s ability to focus on the actual changes in content. In particular, when the style differences between two images are complex and unpredictable, the model may misinterpret unchanged areas—where the only difference lies in style—as changed areas. This misclassification can lead to erroneous results, which undermines the accuracy of the change detection task. Therefore, addressing the interference caused by these inherent style discrepancies is crucial for ensuring that the model focuses solely on the true content changes, which is the primary goal of change detection. 

\begin{figure}[!t]
\centering
\includegraphics[width=3.4in]{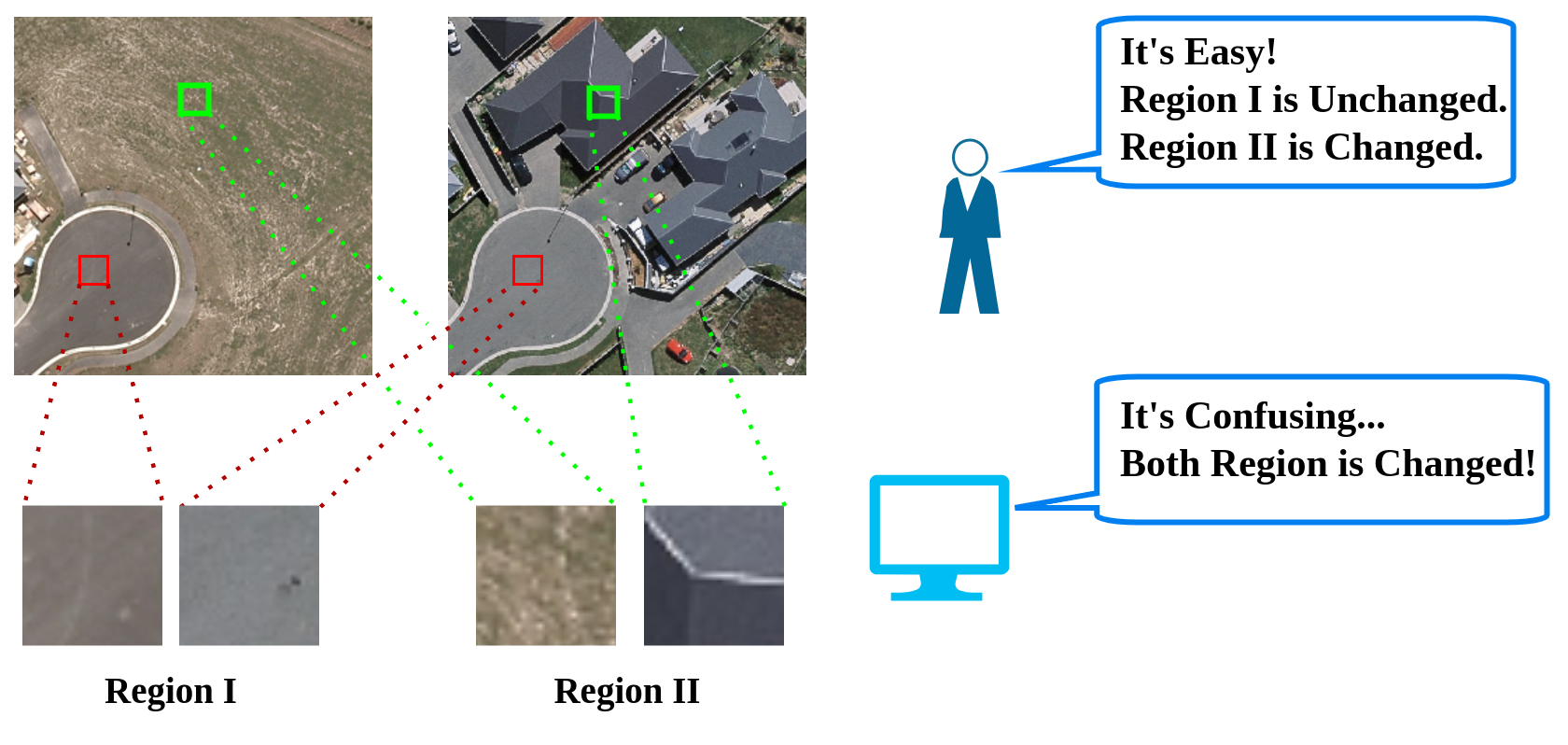}
\caption{humans can easily identify the changed content areas between two images taken under different
conditions, without being significantly influenced by image
style factors such as brightness, contrast, etc. However,this task is sometimes difficult for computer.}
\label{explain}
\end{figure}

Specifically, the definition of ``content" in this paper is based on self-similarity, highlighting that human perception identifies objects by their appearance relative to their surroundings rather than their absolute appearance\cite{kolkin2019style}\cite{gu2018arbitrary}.  Our definition of  ``style" in this paper is a distribution over features extracted by a deep neural network which is mainly caused by imaging conditions including environmental conditions, sensor parameters, acquisition settings, etc\cite{cheng2024harmony}.

\begin{figure*}[!t]
\centering
\includegraphics[width=\textwidth]{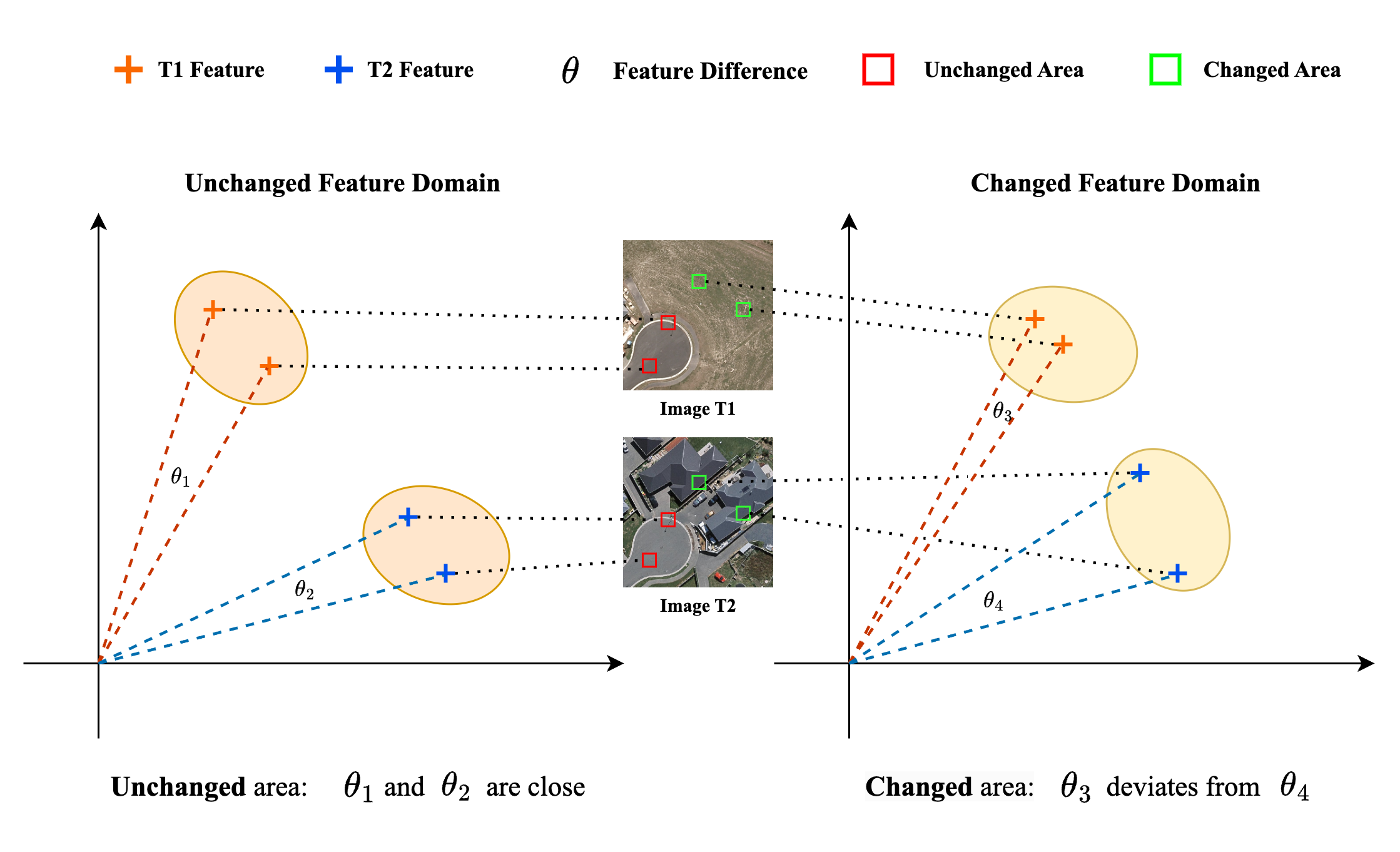}
\caption{ Image T1 and Image T2 exhibit significant style differences. Each ellipse in figure represents the distribution of a pixel's features in feature space. And We use $\theta$ to denote the difference between the features of two sampled points in feature space. Next, we randomly sample two points each from both the changed and unchanged areas. The red boxes in the figure represent sampling points from the unchanged areas, while the green boxes represent those from the changed areas. In the unchanged areas, where the internal structure is similar,the value of $\theta_1$ is close to the value of $\theta_2$.  In contrast, in the changed areas, where the internal structure varies significantly, the value of $\theta_3$ deviates from the value of $\theta_4$. }
\label{fig:content}
\end{figure*}

In recent years, Siamese neural networks have dominated remote sensing change detection by leveraging shared weights for bi-temporal feature extraction\cite{liu2016deep}\cite{daudt2018fully}\cite{li2020siamese}. However, disentangling content and style remains a challenge due to style discrepancies from varying imaging conditions (e.g., lighting, atmosphere, sensor calibration). These differences can mislead models, hindering their focus on actual changes. To address this, CiDL employs dual Y-shaped networks with cross-domain translation to suppress style-induced noise\cite{fang2022}, while CCNet introduces a multi-resolution parallel structure and an auxiliary image restoration task to enhance content-style separation\cite{cheng2024harmony}. Yet, these methods overlook the structural consistency of unchanged areas, a key cue for accurate detection. Moreover, the lack of mechanisms to dynamically balance attention between changed and unchanged areas limits their adaptability in complex scenarios.

As shown in Fig.\ref{explain}, humans can easily identify the content changed areas between two images taken under different conditions, without being significantly influenced by image style factors such as brightness, contrast, etc. This is because when determining whether a area has changed, humans tend to focus more on the internal structure of the area rather than its style features. In fact, this aligns with the cognitive mechanisms of human perception, where humans are accustomed to paying attention to the internal structural information of an object. Based on the above ideas, we propose that even when complex style differences exist between two bi-temporal remote sensing images, the difference in feature vectors at any two locations within unchanged areas remains similar. Therefore, the differences within the set of feature vector differences across any two positions in the unchanged areas should be as small as possible. This set can represent the internal structural characteristics of unchanged areas, as described earlier. Similarly, in the changed areas, the differences within the set of feature vector differences across any two positions in the changed areas should be as large as possible.

As illustrated in Fig.\ref{fig:content}, Image T1 and Image T2 exhibit significant style differences, and thus, even in the unchanged areas of the two images, the distribution of features in the feature space is inconsistent. To simplify the representation, we use a two-dimensional Cartesian coordinate system to illustrate the distribution of multi-dimensional feature vectors in feature space. In the figure, each ellipse represents the distribution of a pixel's features in feature space. We use $\theta$ to denote the difference between the features of two sampled points in feature space. Next, we randomly sample two points each from both the changed and unchanged areas. The red boxes in the figure represent sampling points from the unchanged areas, while the green boxes represent those from the changed areas. In the unchanged areas, where the internal structure is similar, the value of $\theta_1$ is close to the value of $\theta_2$. In contrast, in the changed areas, where the internal structure varies significantly, the value of $\theta_3$ diverges from the value of $\theta_4$. By modeling these characteristics, we aim to enhance the model's ability to learn content features in both changed and unchanged areas. Additionally, the difference in content features between changed and unchanged areas serves as a mutual constraint, further improving the model's accuracy.

In existing change detection algorithms, the concatenation method is widely used for feature fusion of bi-temporal remote sensing images. The reason this method achieves relatively good performance is that DNN have strong generalization capabilities, allowing them to learn the fitting relationship between fused features and labels\cite{liu2016deep}. However, concatenation does not directly compute the changed areas between two images. Therefore, how to make the model focus more on the changed areas during feature fusion, in order to better align with the labels, remains a crucial challenge. In addition, existing change detection algorithms often pay much more attention to the model's ability to fit the changed areas in the labels. However, the binary classification nature of bi-temporal binary change detection tasks means that the accuracy of fitting unchanged areas also significantly impacts the detection results for changed areas in the final prediction map. Therefore, effectively leveraging the mutual constraints between changed and unchanged areas to further enhance the robustness of the model remains a pressing challenge.In this paper, we creatively designed a plug-and-play Focuser module that allows the model to flexibly focus on both changed and unchanged areas depending on the task's requirements at different stages. Building on this, our designed Focuser module not only enables the model to explicitly focus on the changed areas during feature fusion but also leverages the mutual constraints between changed and unchanged areas to impose enhanced parameter regularization, thereby improving the model's robustness and accuracy.

In summary, the main contributions are as follows:
\begin{enumerate}
    \item We propose a Content-Aware strategy, a novel content-based constraint learning strategy that enhances the model's focus on intrinsic content features while reducing the impact of style variations, thereby improving the accuracy and robustness of bi-temporal change detection in remote sensing imagery.
    \item We introduce a plug-and-play Focuser module, a novel mechanism that dynamically reweights features to focus on both changed and unchanged areas, leveraging their mutual constraints to enhance parameter regularization and improve model accuracy.
\end{enumerate}
\section{Related Works}

\subsection{Deep Learning in Change Detection}

Traditional change detection methods, including random forests \cite{ijgi7100401}, support vector machines (SVMs) \cite{7500102}, Markov random fields (MRFs) \cite{bruzzone2000automatic}, decision tree\cite{xie2019hierarchical} and conditional random fields (CRFs) \cite{zhou2016change, hao2019advanced}, rely heavily on handcrafted features and struggle with complex environmental variations. In contrast, deep learning approaches have demonstrated superior performance by automatically extracting hierarchical feature representations, effectively addressing challenges such as image quality variations, noise, registration errors, and spatial heterogeneity \cite{cheng2024change}.

Convolutional Neural Networks (CNNs) have demonstrated remarkable performance in change detection tasks, primarily due to their powerful feature extraction capabilities and non-linear modeling capacity. Unlike traditional methods, CNNs can automatically learn hierarchical spatial representations, thereby improving change localization and classification accuracy. For instance, Alcantarilla et al. developed a structural change detection system for street-view videos using Fully Convolutional Networks \cite{alcantarilla2018street}. Shao et al. introduced a dual-channel network incorporating edge information to address challenges in heterogeneous satellite and UAV image change detection \cite{shao2021sunet}. Lin et al. proposed P2V-CD, a framework that constructs pseudo video sequences and employs decoupled encoders to enhance temporal information processing, significantly improving detection accuracy \cite{lin2022transition}. More recently, Han et al. introduced CGNet, which integrates change guidance maps and a self-attention module to refine feature representations, improving edge integrity and reducing internal noise in change maps \cite{han2023change}. Additionally, Dong et al. proposed EfficientCD, which utilizes EfficientNet as its backbone and integrates ChangeFPN to enhance multi-scale feature aggregation through progressive upsampling, achieving state-of-the-art performance \cite{dong2024efficientcd}.

Transformer-based approaches have recently gained significant attention in change detection due to their capability to model long-range dependencies and capture spatial-temporal correlations more effectively than CNNs. Chen et al. \cite{chen2021remote} proposed the Bitemporal Image Transformer (BIT), which formulates change detection as a token-based representation learning problem. By encoding bitemporal images into a compact set of semantic tokens and leveraging a Transformer encoder-decoder framework, BIT efficiently models spatial-temporal dependencies while significantly reducing computational costs. Bandara et al. \cite{bandara2022transformer} introduced ChangeFormer, a Siamese network architecture that integrates a hierarchically structured Transformer encoder with an MLP decoder, effectively capturing multi-scale contextual information for improved change detection accuracy. More recently, Dong et al. \cite{dong2024changeclip} proposed ChangeCLIP, which extends the Transformer paradigm by incorporating vision-language pretraining, leveraging the semantic representations of image-text pairs to enhance the robustness of change detection. These advancements demonstrate the increasing impact of Transformer architectures in remote sensing change detection, further expanding the scope beyond traditional CNN-based frameworks.

Given the complementary strengths of CNNs in capturing local spatial features and Transformers in modeling long-range dependencies, hybrid CNN-Transformer architectures have been increasingly explored for change detection. These approaches typically utilize CNNs for low-level feature extraction while leveraging Transformers to enhance global contextual representation. Yuan et al. \cite{yuan2022stransunet} proposed STransUNet, which integrates a UNet-based CNN for hierarchical feature extraction with a Transformer module to capture global dependencies, along with a cross-enhanced adaptive fusion module to refine bitemporal feature representations. Xu et al. \cite{xu2024hybrid} introduced HATNet, a hybrid attention-aware Transformer network that incorporates self-attention and coordinate-attention mechanisms to improve multiscale feature extraction and alignment, ensuring better spatial coherence in change maps. These hybrid models effectively combine the advantages of CNNs and Transformers, demonstrating superior performance in remote sensing change detection by balancing local detail preservation and global feature interaction.

\begin{figure*}[!t]
\centering
\includegraphics[width=\textwidth]{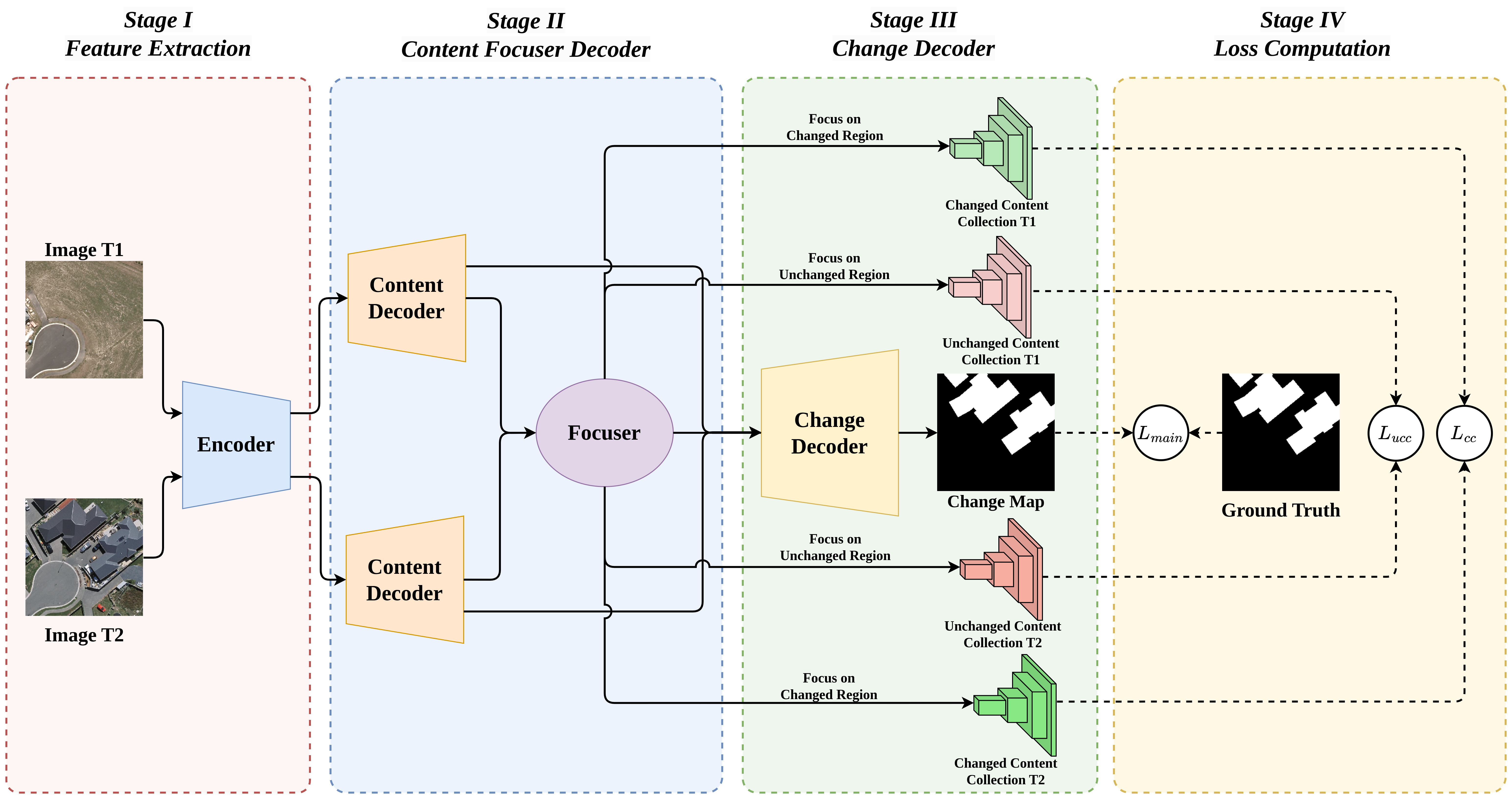}
\caption{The overall architecture of CFNet. The architecture is divided into four key stages: Feature Extraction, Content
Focuser Decoder, Change Decoder, and Loss Computation. In Stage I, a partial EfficientNet-B5 backbone extracts multi-scale features from bi-temporal images. In Stage II, The decoder extracts content features, and the Focuser module generates reweighting maps to separate changed and unchanged content. In Stage III, Content features and reweighting maps are leverged to generate the Change Map. In Stage IV, the total loss consists of the Main Loss  $L_{main}$ , computed using MSE loss between the Change Map and the ground truth, and the auxiliary losses  $L_{ucc}$  and  $L_{cc}$ , which work collaboratively to distinguish changed and unchanged areas, further enhancing the model’s performance. $L_{cc}$ denotes ``Changed Content Loss" and $L_{ucc}$ denotes ``Unchanged Content Loss".}
\label{fig:architecture}
\end{figure*}

\subsection{Separation of Content and Style}
Style transfer refers to the creation of artistic imagery by separating and recombining the content and style of images. Recently, image style transfer has garnered significant attention. In this field, some researchers have proposed methods to effectively separate image style from image content, achieving impressive results.

Gatys et al. illustrated the capability of CNNs to separate and recombine image content and style, pioneering the Neural Style Transfer (NST) technique\cite{Gatys2016}. This method, which applies CNNs to render content images in diverse styles, has since gained significant traction in both academia and industry, inspiring numerous approaches to enhance or expand upon the original NST framework\cite{jing2019neural}. 

Xu et al. introduced a co-analysis method which enables efficient style transfer and synthesis of new 3D objects by defining a correspondence-free style signature for clustering, facilitating part-level correspondence within style clusters\cite{xu2010style}.
Zhange et al. proposed generalized style transfer network enables effective style-content separation using conditional dependence, allowing for versatile style transfer and generalizability across new styles and contents\cite{zhang2018separating}.
Zhange et. al proposed framework that enables a single style transfer model to generalize across multiple styles and contents by leveraging separate style and content encoders, enhancing its applicability to unseen styles and contents\cite{zhang2020unified}.
In 2021, StyleMix and StyleCutMix improve robustness by selectively mixing content and style information in data augmentation, enhancing model generalization and resilience against adversarial attacks\cite{hong2021stylemix}.

Some researchers have creatively identified that the strategy of separating image content and style can be effectively applied to remote sensing change detection tasks.
Fang et al. first introduced the concept of content-style separation into remote sensing change detection tasks\cite{fang2022}. Their proposed CiDL integrates a dual learning algorithm with disentangled representation theory to separate content and style features, suppressing style discrepancies in unchanged areas while highlighting content changes, thereby improving accuracy and reducing dependency on labeled data. Building on this foundation, Cheng et al. further advanced disentangled representation learning by refining the separation of content and style into distinct subspaces, effectively addressing pseudo changes caused by varying imaging conditions and platforms, and enhancing the robustness of change detection frameworks.

In this paper, we extend the concept of content-style separation to the complex task of change detection in remote sensing, with a novel design tailored specifically to enhance the model's ability to focus on the content features of bi-temporal remote sensing images. This approach mitigates the risk of the model being misled by complex style differences between the two temporal images during training.
\section{Method}
\subsection{Overall Architecture}

In this paper, we propose a novel content-based constraint learning strategy, Content-Aware, which focuses on content features. Additionally, we have ingeniously designed a plug-and-play Focuser module that enables the model to flexibly focus on changed and unchanged areas. 

As illustrated in the Fig.\ref{fig:architecture}, the complete architecture is divided into four principle stages: Feature Extraction, Content Focuser Decoder, Change Decoder, and Loss Computation. In the Stage I, Feature Extraction, we employ a part of the EfficientNet-B5 backbone as the encoder to extract rich multi-scale features from the bi-temporal images\cite{tan2019efficientnet}. In TABLE I, the term MBConv refers to a specific module within the architecture. The number following MBConv (e.g., 1 or 6) indicates the expansion factor, which determines how many times the number of input feature matrix channels will be expanded by the first 1x1 convolution layer within the MBConv module. The kernel size represents the size of the convolutional kernel used in the Depthwise Convolution, a critical operation within the MBConv module\cite{chollet2017xception}. The "Resolution" column specifies the input resolution for each stage, while "Channels" refers to the number of output feature matrix channels produced after passing through the stage. Finally, "Layers" indicates the number of times the operator structure is repeated in a given stage. It is worth noting that the bi-temporal images share the same weights during the encoding process. The detailed structure of Encoder is presented in TABLE \ref{tab:Encoder}. The feature maps extracted from stage 2 to stage 5 of the bi-temporal images are used as the output of the Encoder, denoted as $F_{a1}$, $F_{a2}$, $F_{a3}$, $F_{a4}$ for image T1, and $F_{b1}$, $F_{b2}$, $F_{b3}$, $F_{b4}$ for the image T2, respectively. 

\begin{figure}[!t]
\centering
\includegraphics[width=3.4in]{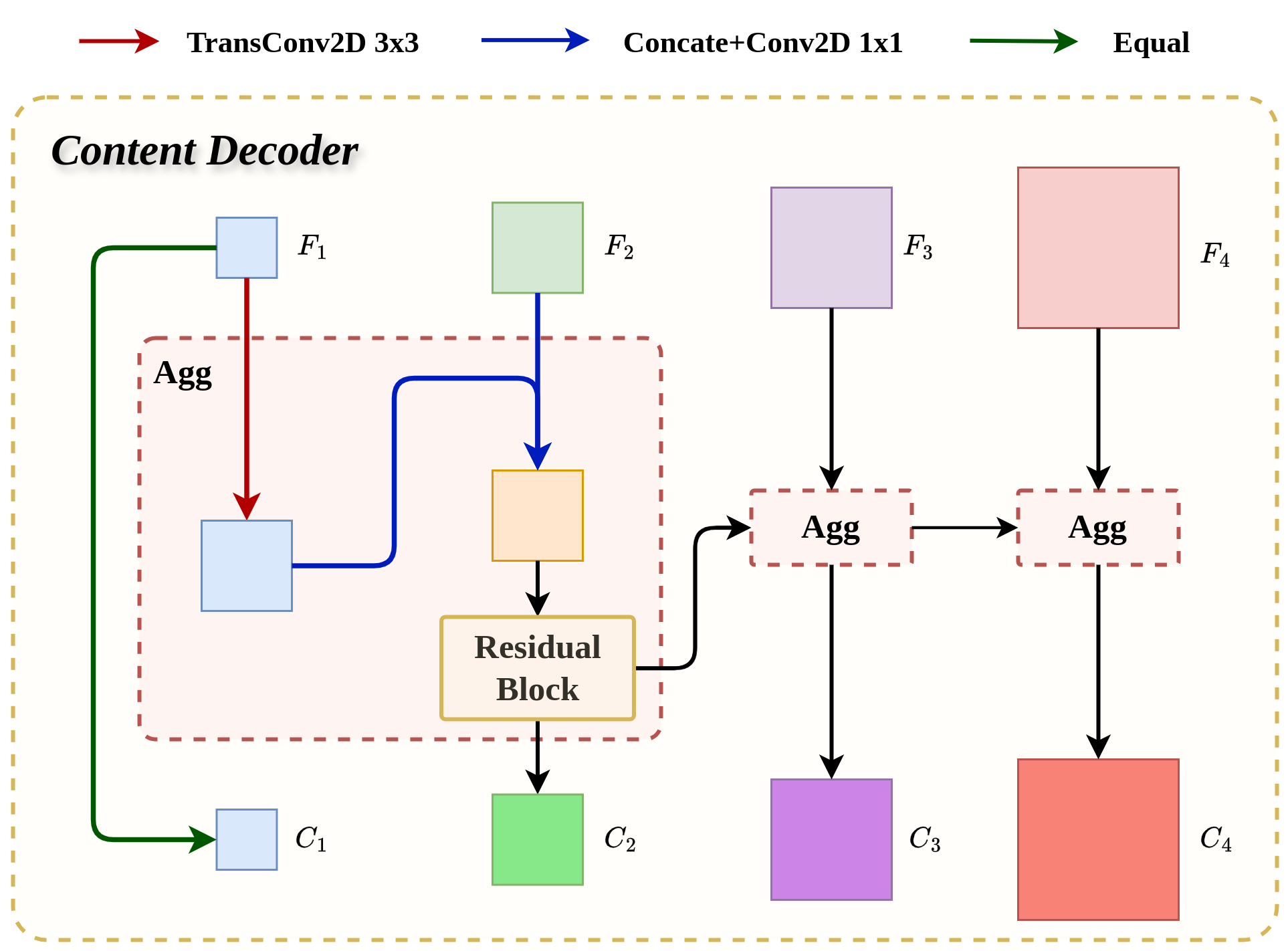}
\caption{The detailed architecture of Content Decoder. $F_{i}, i=1,2,3,4$ denotes the output of the Encoder, specifically $F_{ai}$ or $F_{bi}$. $C_{i}, i=1,2,3,4$ represents content feature, specifically $C_{ai}$ or $C_{bi}$. The Agg module is used to aggregate the feature maps from adjacent scales of the encoder's output.}
\label{fig:contentdecoder}
\end{figure}

In the Stage II, the Content Focuser Decoder, the model first decodes the multi-scale features extracted from the bi-temporal images in the Content Decoder module to obtain content features at different scales, as shown in Fig.\ref{fig:contentdecoder}.
\begin{table}
\begin{center}
\caption{THE DETAILED ARCHITECTURE OF ENCODER.}
\label{tab:Encoder}
\setlength{\arrayrulewidth}{0.4mm}
\renewcommand{\arraystretch}{1.3} 
\begin{tabular}{ c c c c c c }
\hline
Stage & Operator& Kernel Size & Resolution & Channels & Layers\\
\hline
1 & Conv & 3x3& 128x128 &48 & 1 \\
2 & MBConv1 &3x3& 128x128 &24 & 3 \\
3 & MBConv6 & 3x3& 64x64 &40 & 5 \\
4 & MBConv6 & 5x5& 32x32 &64 & 5 \\
5 & MBConv6 & 3x3& 16x16 &128 & 7 \\
\hline
\end{tabular}
\end{center}
\end{table}
It is important to clarify that in Fig.\ref{fig:contentdecoder}, $F_{i}, i=1,2,3,4$ denotes the output of the Encoder, specifically $F_{ai}$ or $F_{bi}$, while $C_{i}, i=1,2,3,4$ represents content feature, specifically $C_{ai}$ or $C_{bi}$ . Although the two Change Decoder modules share the same architecture, they operate with independent weights. The Agg module is used to aggregate the feature maps from adjacent scales of the encoder's output. The multi-scale content features are subsequently passed through the Focuser module, which generates reweighting maps corresponding to each scale. These reweighting maps, denoted as $RM_{i},i=1,2,3,4$,  retain the same scale as their respective input content features, ensuring alignment across scales. Subsequently, the Focuser module re-weights the changed and unchanged areas of the bi-temporal images, effectively separating changed content features from unchanged content features. This results in the ``Changed Content Collection" and ``Unchanged Content Collection" for the bi-temporal images. The structure and function of the Focuser module will be discussed in detail in Section \ref{sec:Focuser}. 

\begin{figure}[!t]
\centering
\includegraphics[width=3.45in]{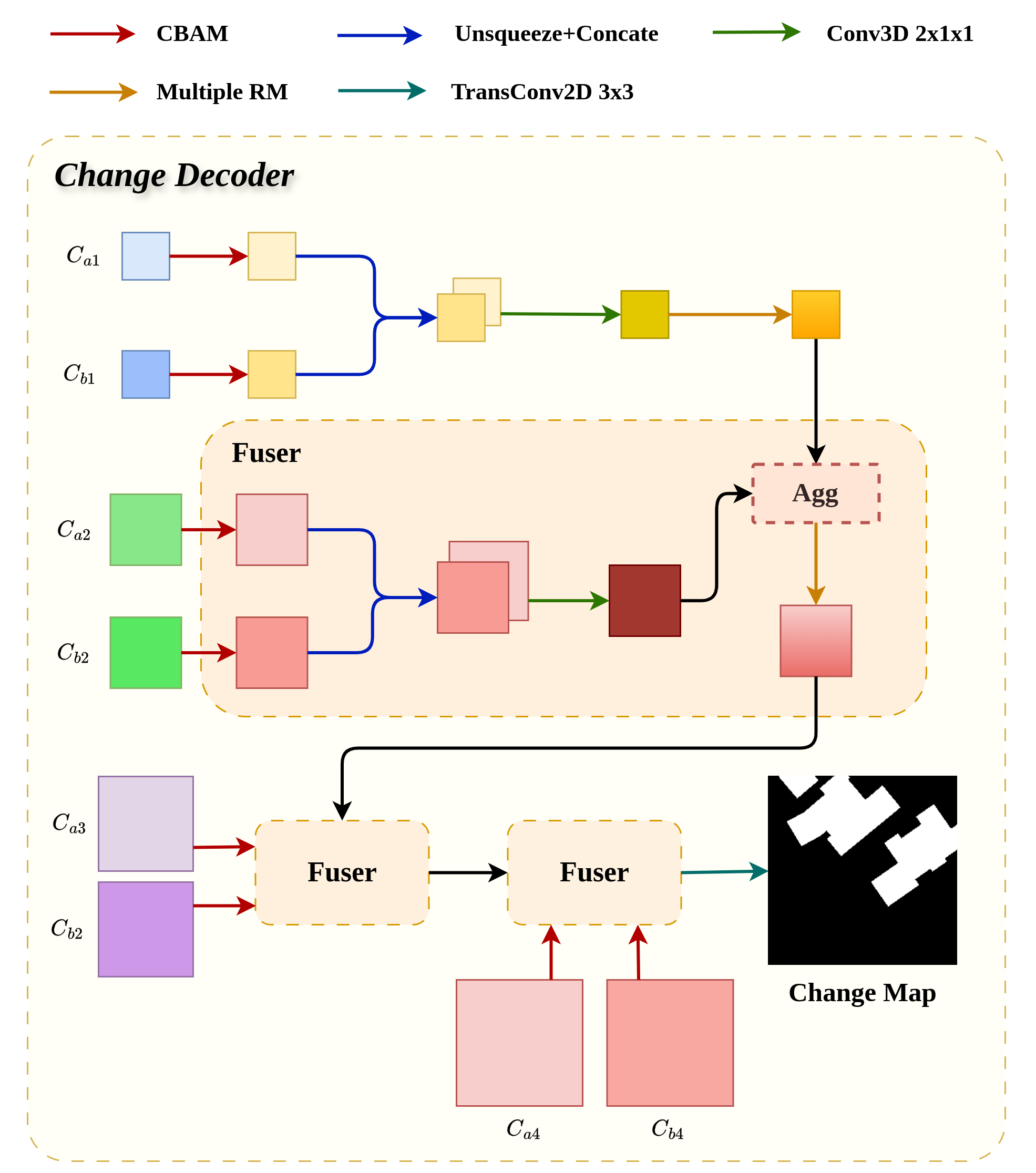}
\caption{The detailed architecture of Change Decoder. The red arrows labeled as ``CBAM" represents Convolutional Block Attention Module\cite{woo2018cbam}.It is a lightweight module that enhances feature representation by applying channel and spatial attention, helping the network focus on the most relevant features and areas. The blue arrows labeled "Unsqueeze+Concate" indicates that two inputs are each expanded by an identical new dimension, concatenated along this new dimension to produce an output, which is then used for subsequent 3D convolution operations. The yellow arrows labeled as ``Multiple RM" represents performing a dot product between the feature map at the starting point of the arrow and the corresponding scale $RM_{i}$ from Focuser module.} 
\label{fig:changedecoder}
\end{figure}

In the Stage III, Change Decoder, the content features at each scale are fully fused, and the reweighting maps generated in the Stage II are utilized at each decoding stage to focus on the changed areas, eventually producing the Change Map. The detailed structure of Change Decoder is illustrated as Fig.\ref{fig:changedecoder}. Additionally, the specific structure of the Agg module is detailed in Fig.\ref{fig:contentdecoder}. 

In the Stage IV, Loss Computation, the Change Map is compared with the Ground Truth to compute the Main Loss, which is calculated using MSE loss. At the same time, ``Changed Content Collection" and ``Unchanged Content Collection" from the Stage II leveraged for self-supervision, enabling the computation of the ``Changed Content Loss" and ``Unchanged Content Loss". This computation is central to the Content-Aware strategy, which will be discussed in detail in Section \ref{sec:content}.

\begin{figure*}[!t]
\centering
\includegraphics[width=5.5in]{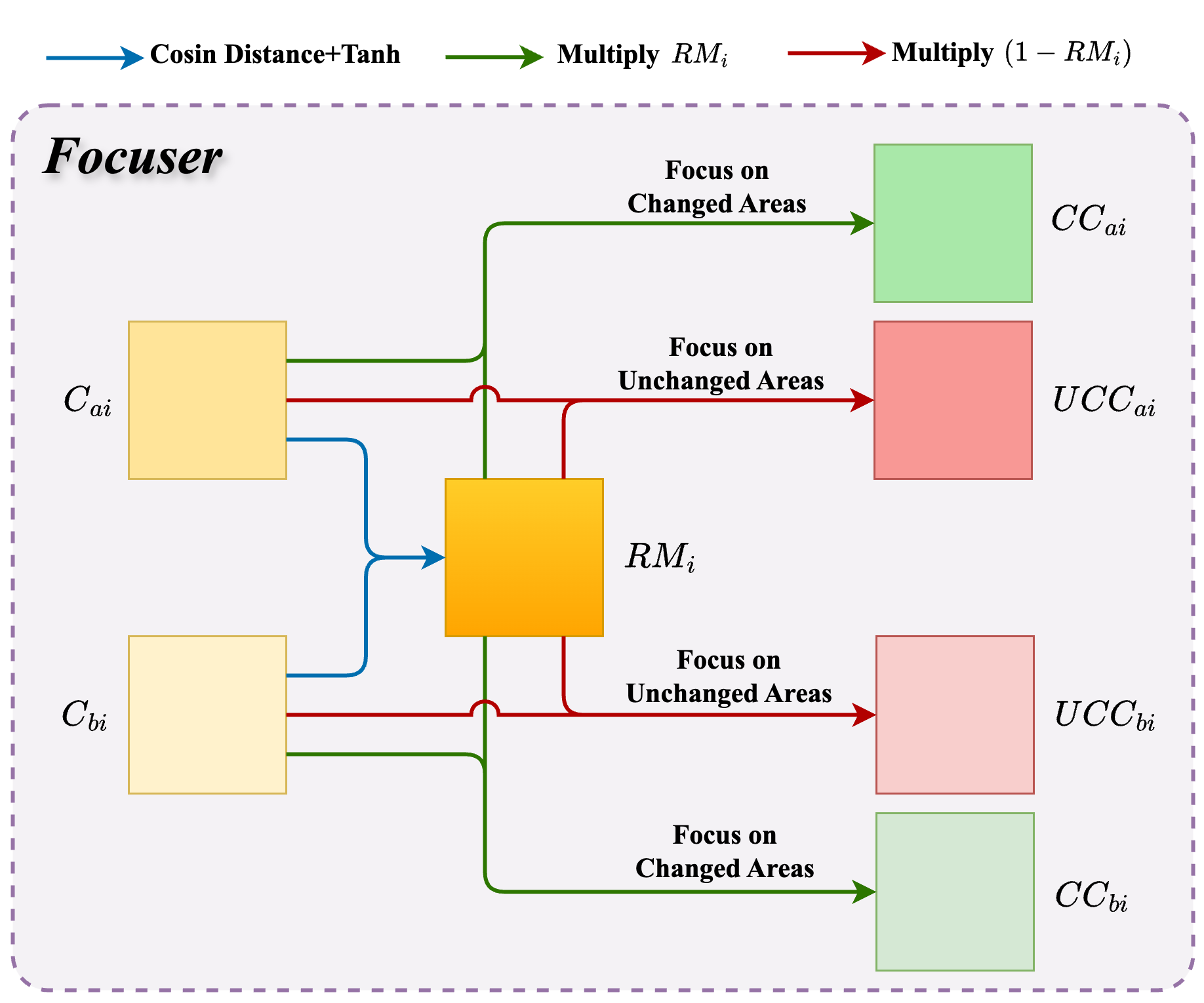}
\caption{The detailed architecture of Focuser module. $C_{ai}$ and $C_{bi}$ represent the content features of bi-temporal images T1 and T2 at a certain scale, as output by the Content Decoder module.$RM_{i}$ reflects the distribution of changed and unchanged areas, where pixels with values closer to 1 are more likely to belong to changed areas, and pixels with values closer to 0 are more likely to belong to unchanged areas. Moreover, ``CC" denotes changed content features and ``UCC" denotes unchanged content features.}
\label{fig:focuser}
\end{figure*}

\subsection{Focuser Module}
\label{sec:Focuser}
In previous research, researchers have typically chosen to concatenate feature maps at multiple scales and then leverage the large number of model parameters to fit the relationships with changed areas. However, this approach, to some extent, overlooks the essence of the change detection task. We argue that the essence of bi-temporal binary change detection is a binary classification task with two inputs. More specifically, the task involves predicting whether a pixel belongs to a changed area or an unchanged area based on bi-temporal remote sensing images. Therefore, we believe the model should flexibly focus on features within the changed area and unchanged area separately during the decoding stage, thereby reducing the reliance on strong model fitting ability. This approach also enables the model to learn features from the changed area and unchanged area independently. Based with this idea, we propose a plug-and-play Focuser module, with its detailed structure shown in Fig.\ref{fig:focuser}. It is important to note that the Focuser module performs identical computations on the content features of bi-temporal remote sensing images at different scales, outputted by the Content Decoder. Therefore, in the figure, we use $C_{ai}$ and $C_{bi}$ to represent the content features of bi-temporal images at the same scale, illustrating the consistent computation process of the Focuser module across all scales.

As shown in Fig.\ref{fig:focuser}, $C_{ai}$ and $C_{bi}$ represent the content features of bi-temporal images T1 and T2 at a certain scale, as output by the Content Decoder module. First, we calculate the cosine distance between $C_{ai}$ and $C_{bi}$ at corresponding positions to quantify the degree of content feature differences at each location. The reason for using cosine distance here is that it helps mitigate the influence of absolute feature magnitude differences, thereby reducing the interference from style variations, which allows us to focus more on the content differences. Next, we apply the Tanh function to normalize the cosine distance map between $C_{ai}$ and $C_{bi}$, generating the Reweighting Map $RM_{i}$ for this scale. $RM_{i}$ reflects the distribution of changed and unchanged areas, where pixels with values closer to 1 are more likely to belong to changed areas, and pixels with values closer to 0 are more likely to belong to unchanged areas. Consequently, in $1-RM_{i}$, pixels with values closer to 1 are more likely to be unchanged areas, while values closer to 0 indicate changed areas. Finally, we multiply $C_{ai}$ and $C_{bi}$ by $RM_{i}$ and $1-RM_{i}$ , respectively, to obtain $CC_{ai}$, $CC_{bi}$, $UCC_{ai}$, and $UCC_{bi}$, where ``CC" denotes change content features and ``UCC" denotes unchanged content features. This approach enables the model to focus on learning both changed and unchanged areas separately during training, thereby improving its performance in change detection task. The computational process mentioned above is presented in the following formulas:

\begin{equation}
\label{eq:0}
\text{Cosine Similarity}=\frac{C_{ai} \cdot C_{bi}}{\|C_{ai}\| \|C_{bi}\|}
\end{equation}

\begin{equation}
\label{eq:1}
\text{Cosine Distance Map}=\displaystyle\sum_{Channel} {1-\text{Cosine Similarity}}
\end{equation}

\begin{equation}
\label{eq:2}
RM_{i} =\tanh (\text{Cosine Distance Map})
\end{equation}

\begin{equation}
\label{eq:3}
CC_{ai}=RM_{i} \cdot C_{ai}
\end{equation}

\begin{equation}
\label{eq:4}
CC_{bi}=RM_{i} \cdot C_{bi}
\end{equation}

\begin{equation}
\label{eq:5}
UCC_{ai}=(1-RM_{i}) \cdot C_{ai}
\end{equation}

\begin{equation}
\label{eq:6}
UCC_{bi}=(1-RM_{i}) \cdot C_{bi}
\end{equation}

\subsection{Content-Aware Strategy}
\label{sec:content}
Due to varying imaging conditions, there are often complex and unpredictable style differences between bi-temporal remote sensing images. In change detection task, the primary focus should be on meaningful content changes, but discrepancies in image style often mislead the model's judgment. We aim to guide the model to learn content features from both changed and unchanged areas through an auxiliary loss function, thereby constraining the network's fitting.

As mentioned in the introduction, when humans assess whether a specific area in an image has changed, they can easily discount the interference caused by style differences. This is because human perception of an object is not solely dependent on individual pixels, but rather on the internal structural features of the object. Inspired by self-similarity, the internal structural features referenced here can be mathematically represented by the set of cosine similarities between feature vectors at any two locations within a area. The use of cosine similarity is motivated by its ability to reduce the influence of feature vector magnitude, thereby better mitigating the interference from style differences.

\begin{figure*}[!t]
\centering
\includegraphics[width=6in]{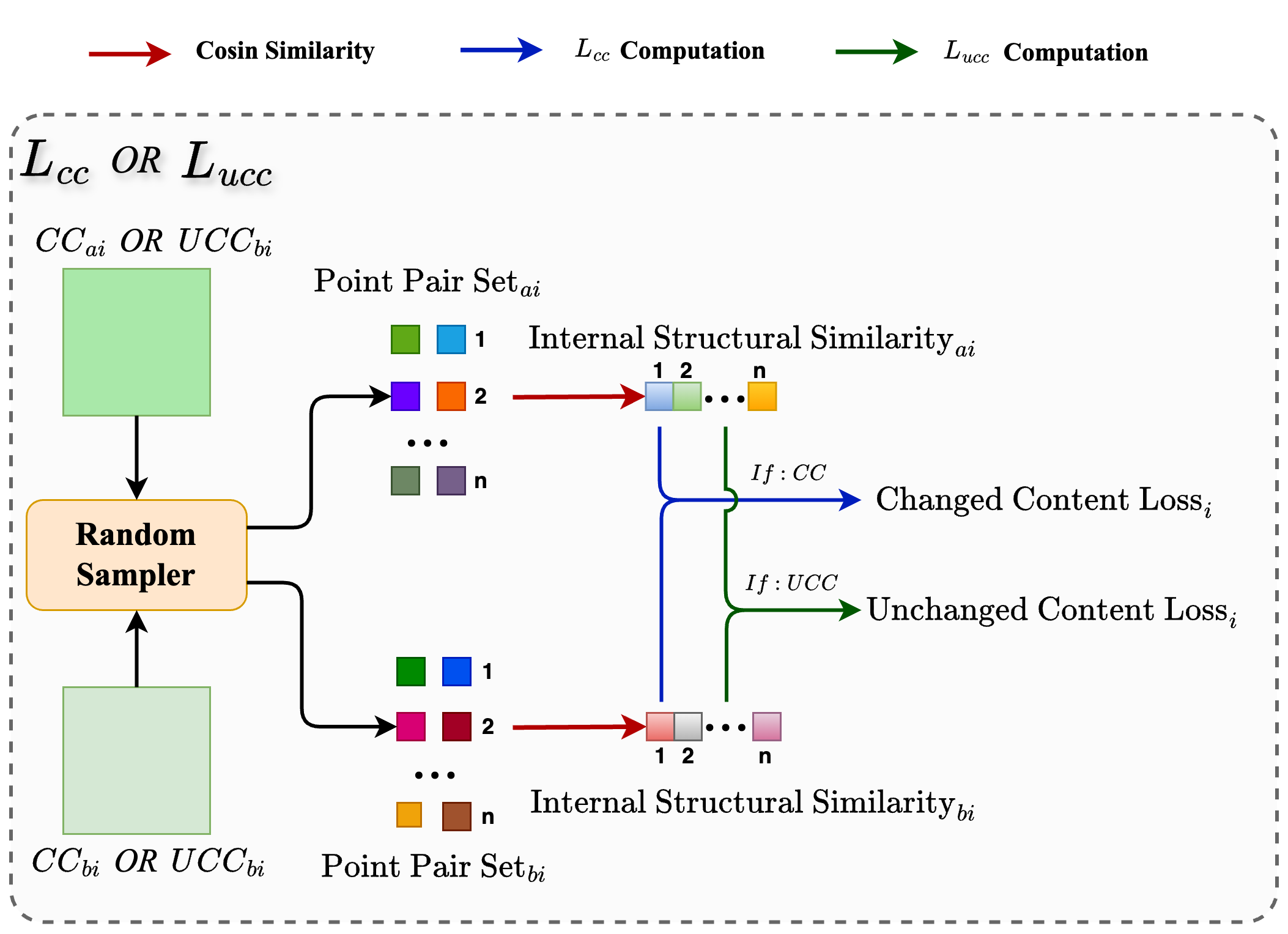}
\caption{This figure demonstrate the detailed calculation process of $L_{cc}$ and $L_{ucc}$.We leverage the Random Sampler to randomly sample $n$ different points from the corresponding positions in $CC_{ai}$,$CC_{bi}$ or $UCC_{ai}$,$UCC_{bi}$, resulting in the $\text{Point Pair Set}_{ai}$ and $\text{Point Pair Set}_{bi}$. For each of these sets, we compute the pairwise cosine similarity between the points, yielding $1\times n$ matrices $\text{Internal Structural Similarity}_{ai}$ and $\text{Internal Structural Similarity}_{bi}$. If the input consists of \(CC_{ai}\) and \(CC_{bi}\), we calculate \(\text{Changed Content Loss}_i\). Conversely, if the input consists of \(UCC_{ai}\) and \(UCC_{bi}\), we compute \(\text{Unchanged Content Loss}_i\).}

\label{fig:content_loss}
\end{figure*}

 Building on the aforementioned considerations, we propose a novel content-based constraint strategy focused on image content features, named Content-Aware. We compute the ``Changed Content Loss"  and ``Unchanged Content Loss" at each scale using the ``Changed Content Collection" and “Unchanged Content Collection” for both Image T1 and Image T2. As a key component of the Content-Aware strategy, the detailed computation process is illustrated in Fig.\ref{fig:content_loss}. Since the calculation steps for ``Changed Content Loss" and ``Unchanged Content Loss" differ only in the final step, we explain the computation of ``Changed Content Loss" first, followed by the specific details of the final step for ``Unchanged Content Loss".

To ensure the computational efficiency of the model, we performed \(n\) rounds of random sampling. In each sampling round, we randomly selected the same two locations from \(CC_{ai}\) and \(CC_{bi}\). The two sets of sampled point pairs were then added as new elements to \(\text{Point Pair Set}_{ai}\) and \(\text{Point Pair Set}_{bi}\), respectively. As a result, both \(\text{Point Pair Set}_{ai}\) and \(\text{Point Pair Set}_{bi}\) ultimately contained \(n\) pairs of sampled points. Subsequently, we calculated the cosine similarity of the feature vectors corresponding to each pair of sampled points in \(\text{Point Pair Set}_{ai}\) and \(\text{Point Pair Set}_{bi}\), yielding two \(1 \times n\) matrices, \(\text{Internal Structural Similarity}_{ai}\) and \(\text{Internal Structural Similarity}_{bi}\).
 In the formulas, we denote Internal Structural Similarity as ``ISS". These matrices represent subsets of internal structural features, each containing $n$ elements. Additionally, to balance computational efficiency and accuracy, in this experiment, we set the number of sampled point pairs \(n\) to be the square root of the scale of \(CC_{ai}\) and \(CC_{bi}\). For example, if the scale of \(CC_{ai}\) is \(256 \times 256\), then \(n\) is set to 256.Since the sampling process is random in each training iteration, we consider these subsets to approximate the overall internal structural features. In essence, the ``Changed Content Loss" should diverge as far from zero as possible, and it is computed according to formula \ref{eq:7}. In contrast, the ``Unchanged Content Loss" should approach zero, and it is computed using formula \ref{eq:8}. Finally, $L_{cc}$ is computed according to formula \ref{eq:9}, while $L_{ucc}$ is calculated as formula \ref{eq:10}. Additionally, it is important to note that to minimize the disparity between $L_{m}$, $L_{cc}$, and $L_{ucc}$, thereby preventing any single loss function from dominating the model's learning process, we applied appropriate weighting to $L_{m}$, $L_{cc}$, and $L_{ucc}$. The final loss computation is presented in formula \ref{eq:11}. During the experimental process, we set \(\alpha\) to 1, \(\beta\) to 0.1, and \(\gamma\) to 0.1.

\begin{equation}
\label{eq:7}
\text{Changed Content Loss}_{i}=1-\frac{1}{n}\sum  |ISS_{ai}- ISS_{bi}|
\end{equation}

\begin{equation}
\label{eq:8}
\text{Unchanged Content Loss}_{i}=\frac{1}{n}\sum  |ISS_{ai}- ISS_{bi}|
\end{equation}

\begin{equation}
\label{eq:9}
L_{cc}=\frac{1}{4}\sum\limits_{i=1}^{4}{\text{Changed Content Loss}_{i}}
\end{equation}

\begin{equation}
\label{eq:10}
L_{ucc}=\frac{1}{4}\sum\limits_{i=1}^{4}{\text{Unchanged Content Loss}_{i}}
\end{equation}

\begin{equation}
\label{eq:11}
Loss=\alpha L_{main}+\beta L_{cc}+\gamma L_{ucc}
\end{equation}

\section{Experiments}
\subsection{Datasets}
To validate the effectiveness and robustness of the proposed CFNet, we conducted experiments on three publicly available and well-known remote sensing change detection datasets: CLCD \cite{liu2022cnn}, LEVIR-CD \cite{chen2020spatial}, and SYSU-CD \cite{shi2021deeply}.
The CLCD dataset is selected for its agriculture-related change detection focus and the non-fixed style differences between bi-temporal images, which allow for evaluating the model's robustness under complex style variations.The LEVIR-CD dataset is chosen for its suitability in building change detection tasks, particularly for evaluating performance on very high-resolution (VHR) images and analyzing fine-grained building additions and demolitions. The SYSU-CD dataset is used to assess the adaptability of methods across various scenarios, given its focus on localized urban change detection and micro-level urban structural changes. It is worth noting that, due to the complexity of imaging conditions, all three datasets feature bi-temporal images with unquantifiable and unpredictable style differences. These inherent style variations present an additional challenge and provide a more comprehensive test for the robustness of the proposed CFNet.

\textbf{CLCD}: The CLCD dataset contains 600 pairs of cropland change samples, divided into 360 pairs for training, 120 pairs for validation, and 120 pairs for testing. The bi-temporal images in the CLCD dataset were captured by the Gaofen-2 satellite in Guangdong Province, China, in 2017 and 2019, respectively, with spatial resolutions ranging from 0.5 to 2 meters. Each sample pair includes two 512 × 512 images along with a corresponding binary cropland change label. During training phase, each 512 x 512 image is divided into four 256 x 256 patches without overlapping. For testing, we use the original 512x512 images from the test set in the dataset.

\textbf{LEVIR-CD}: LEVIR-CD comprises 637 pairs of very high-resolution (VHR, 0.5 m/pixel) Google Earth (GE) image patches, each sized 1024 × 1024 pixels. These bi-temporal images span a period of 5 to 14 years and capture significant land-use changes, particularly in construction growth. The dataset includes a wide variety of building types, such as villas, tall apartment buildings, small garages, and large warehouses. The primary focus is on building-related changes, including building growth (e.g., transitions from soil, grass, or construction sites to fully developed structures) and building decline. All bi-temporal images are annotated by remote sensing experts with binary labels (1 for the changed, 0 for the unchanged). To ensure annotation quality, each sample is first labeled by one annotator and then reviewed by a second annotator. The fully annotated dataset includes 31,333 individual instances of building changes. During training phase, each 1024 x 1024 image is divided into multiple 256x256 patches, with a 64-pixel overlap between adjacent patches. Using this method, a single 1024x1024 image is cut into 25 patches of size 256x256. As a result, we generate 11,125 image pairs for the training set and 1,600 image pairs for the validation set. For testing, we use the original 128 image pairs with a size of 1024x1024 from the dataset.

\textbf{SYSU-CD}: The SYSU-CD dataset consists of 20,000 pairs of high-resolution aerial images, each with a resolution of 0.5 meters and dimensions of 256 × 256 pixels. Captured in Hong Kong between 2007 and 2014, this dataset represents a diverse array of urban changes, including the construction of new buildings, suburban sprawl, groundwork prior to construction, alterations in vegetation, road expansions, and offshore construction projects. The dataset is systematically divided into three subsets: the training set containing 12,000 image pairs, the validation set with 4,000 pairs, and the test set also comprising 4,000 pairs. This organization adheres to widely accepted experimental protocols, making SYSU-CD a valuable resource for assessing and comparing the performance of change detection algorithms across various urban scenarios. Each image pair is annotated with binary labels, indicating whether the pixels have changed or remained unchanged, further enhancing its utility for research in change detection.

\subsection{Benchmark Methods}
To demonstrate the superiority of the proposed CFNet, we selected state-of-the-art algorithms for remote sensing change detection and conducted a comparative evaluation of CFNet across three datasets: CLCD, LEVIR-CD, and SYSU-CD. These algorithms are categorized into three groups based on their architectural paradigms: CNN-based methods, which leverage convolutional operations for feature extraction and spatial pattern recognition; Transformer-based methods, which model long-range dependencies through self-attention mechanisms; and Hybrid CNN-Transformer methods, which combine the strengths of both architectures to enhance spatial and contextual feature representations.

\textbf{CNN-based}:
CDNet\cite{alcantarilla2018street}, SNUNet\cite{9355573}, DDCNN\cite{peng2020optical}, STANet\cite{chen2020spatial}, DDCNN\cite{peng2020optical}, P2V\cite{lin2022transition}, HCGMNet\cite{han2023hcgmnet}, CGNet\cite{han2023change}, AFCF3D-Net\cite{ye2023adjacent}, ChangeEx\cite{fang2023changer}, EfficientCD\cite{dong2024efficientcd}, CDNeXt\cite{wei2024robust}.

\textbf{Transformer-based}:
BIT\cite{chen2021remote}, ChangeFormer\cite{bandara2022transformer}, AMTNet\cite{liu2023attention}, ChangeCLIP\cite{dong2024changeclip},.

\textbf{Hybrid CNN-Transformer}:
  DMATNet\cite{song2022remote}, DARNet\cite{li2022densely}, SSANet\cite{jiang2022joint}, STransUNet\cite{yuan2022stransunet}, MSCANet\cite{liu2022cnn}, DMINet\cite{feng2023change}, GAS-Net\cite{zhang2023global}, HATNet\cite{xu2024hybrid},  .

\begin{table}
\begin{center}
\caption{QUANTITATIVE RESULTS ON CLCD}
\label{tab:clcd_res}
\setlength{\arrayrulewidth}{0.4mm}
\renewcommand{\arraystretch}{1.3} 
\resizebox{3.4in}{!}{
\begin{tabular}{ c  c  c  c  c }
\hline
Model & F1 & IoU & Rec & Prec\\
\hline
CDNet & 65.15 &48.31 &59.43 &72.09 \\
STANet& 67.97 & 51.49& 64.16& 72.26\\
P2V&70.22 &54.10 &65.93 & 75.11\\
MSCANet& 71.65& 55.83& 67.07 & 76.91\\
HATNet& 72.53& 56.90& 69.42& 75.94\\
BIT &73.3 &57.85 &71.84 & 74.82\\
HCGMNet& 74.67&59.58 &73.92&75.44 \\
DMINet& 76.51&61.96 &71.98& 81.65 \\
AMTNet& 76.81& 62.35& 75.06& 78.64\\
CGNet& 77.05 & 62.67& 71.71& \textbf{83.25}\\
EfficientCD& 78.89 & 65.14& 75.83& 82.21\\
\textbf{CFNet}& \textbf{81.41} & \textbf{68.65} & \textbf{81.08} & 81.74\\
\hline
\end{tabular}
}
\end{center}
\end{table}

\begin{figure*}[!t]
\centering
\includegraphics[width=\textwidth]{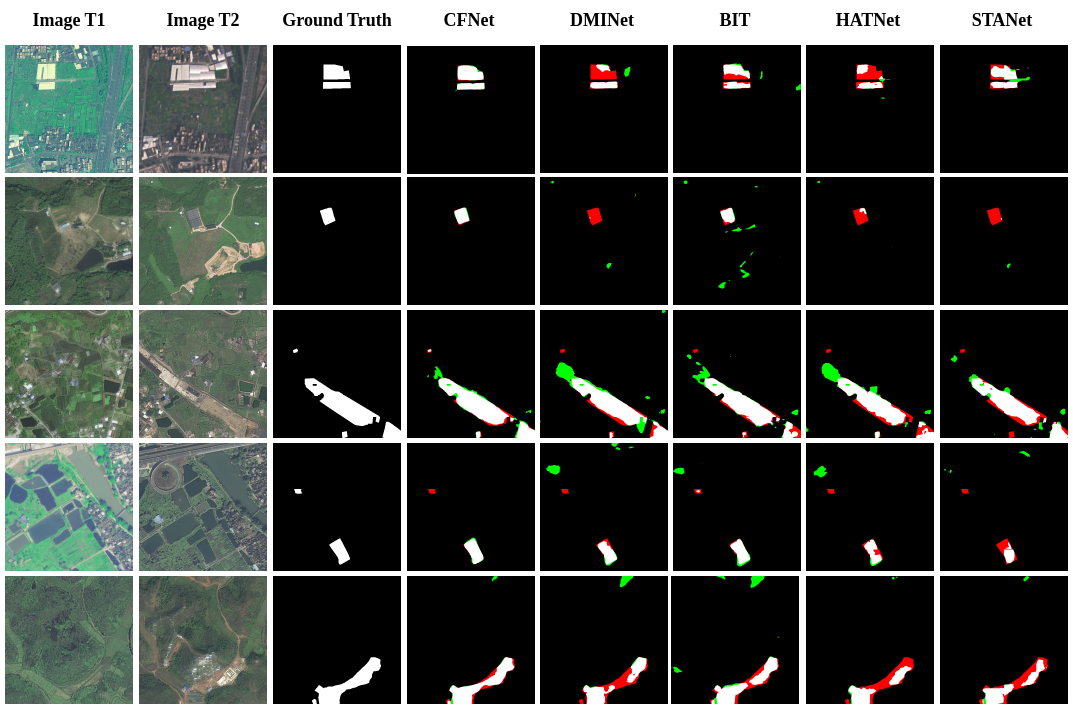}
\caption{The visualization result in CLCD dataset.}
\label{fig:cl_vis}
\end{figure*}

\subsection{Experimental Detail}
In this study, we implemented the proposed CFNet using the PyTorch framework. The experiments were conducted on a machine with Ubuntu 24.04 LTS as the operating system. The GPU used was an NVIDIA GeForce RTX 3090, while the CPU was an Intel(R) Xeon(R) Gold 6130H CPU @ 2.10GHz with 64 cores. For data augmentation, we applied random flipping, scaling, translation, rotation, along with the addition of Gaussian blur and contrast adjustment. Regarding training specifics, we utilized a single NVIDIA GeForce RTX 3090 GPU throughout the experiments, setting the batch size to 32. The AdamW optimizer was employed to optimize the training process, with an initial learning rate of 0.0005, dynamically adjusted using a cosine annealing scheduler. During training, we continuously monitored the model's Intersection over Union (IoU) and F1 score on the validation set. Whenever there was an improvement in either IoU or F1 compared to previous iterations, we saved the current model weights and subsequently evaluated the model's performance on the test set.

\subsection{Evaluation Metrics}
To evaluate the effectiveness of our model, we focused on four key metrics: Intersection over Union (IoU), F1 score, recall, and precision. IoU is vital for assessing the accuracy of change detection by measuring the overlap between predicted and actual changes. Precision reduces false positives, ensuring that the detected changes are relevant, while recall minimizes missed detections, capturing all relevant changes. The F1 score balances precision and recall, providing a comprehensive view of model performance, especially in scenarios with class imbalances. These metrics were selected for their ability to offer a detailed understanding of both overall and individual performance, crucial for assessing our algorithm's effectiveness.  The calculation formulas are as follows:
\begin{equation}
\label{deqn_ex1}
\text{IoU} = \frac{TP}{TP + FP + FN}
\end{equation}

\begin{equation}
\label{deqn_ex1}
\text{Recall} = \frac{TP}{TP + FN}
\end{equation}

\begin{equation}
\label{deqn_ex1}
\text{Precision} = \frac{TP}{TP + FP}
\end{equation}

\begin{equation}
\label{deqn_ex1}
F1 = 2 \cdot \frac{\text{Precision} \cdot \text{Recall}}{\text{Precision} + \text{Recall}}
\end{equation}

True Positives (TP) refer to correctly identified positive pixels, indicating that the model accurately detects changes. False Positives (FP) represent incorrectly identified positives, where the model mistakenly detects changes that aren't present. False Negatives (FN) are missed positives, meaning the model fails to detect actual changes, while True Negatives (TN) denote correctly identified negatives, where the model accurately identifies areas without changes. Together, these terms are essential for evaluating model performance in classification tasks, particularly in change detection task.

\begin{figure*}[!t]
\centering
\includegraphics[width=\textwidth]{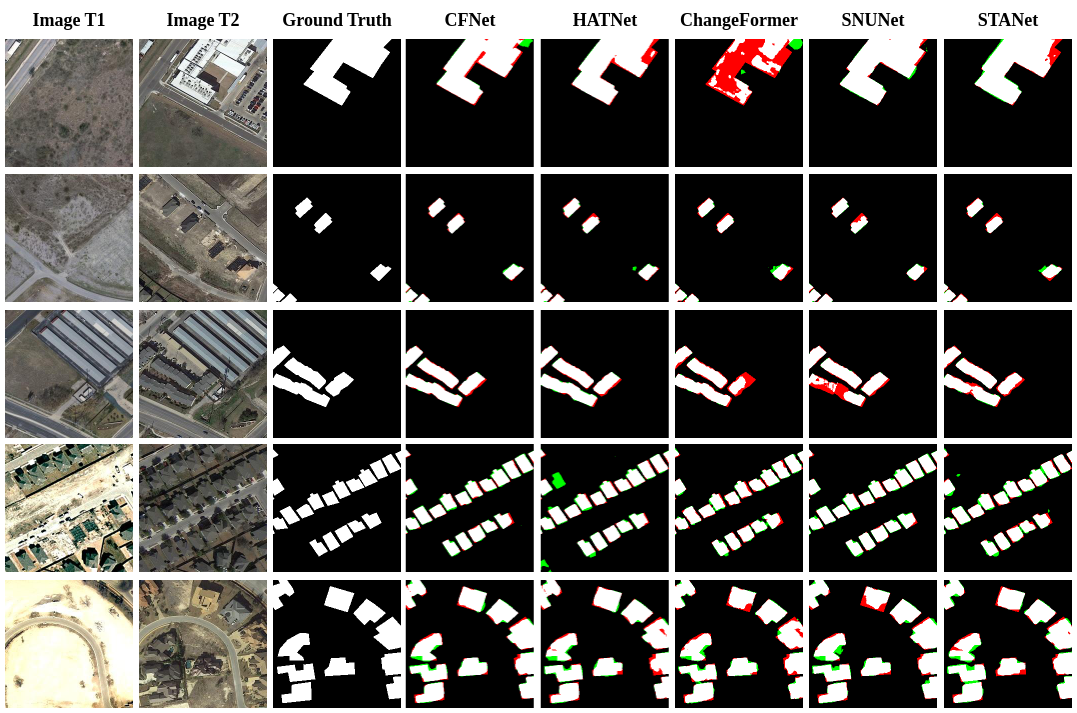}
\caption{The visualization result in LEVIR-CD dataset.}
\label{fig:levir_vis}
\end{figure*}

\subsection{Qualitative Analysis and Visualization}
As shown in TABLE \ref{tab:clcd_res}, TABLE \ref{tab:levir} and TABLE \ref{tab:sysu}, we validated the proposed CFNet on three well-known public remote sensing change detection datasets: CLCD, LEVIR-CD, and SYSU-CD. We compared the performance metrics of our experimental results with several state-of-the-art algorithms in remote sensing change detection, demonstrating that CFNet achieves state-of-the-art levels for each dataset.

\begin{table}
\begin{center}
\caption{QUANTITATIVE RESULTS ON LEVIR-CD}
\label{tab:levir}
\setlength{\arrayrulewidth}{0.4mm}
\renewcommand{\arraystretch}{1.3} 
\resizebox{3.3in}{!}{
\begin{tabular}{ c  c  c  c  c }
\hline
Model & F1 & IoU & Rec & Prec\\
\hline
BIT& 89.48 & 80.86 & 87.53 & 90.65 \\
STANet &90.02&81.85&87.13&93.10\\
MSCANet& 90.06& 81.91& 86.38& 94.06 \\
DDCNN& 90.24& 82.21&88.69 & 91.85\\
SNUNet& 90.33& 82.36&87.62 & 93.2\\
ChangeFormer&90.50 &82.66 &90.18 &90.83 \\
P2V&90.71 &83.00 &91.78 &89.67 \\
DMATNet&90.75 &84.13 &89.98 &91.56 \\
HATNet & 91.38&84.12&90.15&92.64\\
STransUNet& 91.41&84.19 &90.55 &92.30 \\
ChangeCLIP& 92.01&85.20 &90.67 &93.40 \\
ChangerEx& 92.06&85.29 & 90.56& \textbf{93.61}\\
CGNet& 92.13&85.40 & \textbf{91.93}& 92.32\\
\textbf{CFNet}& \textbf{92.18} & \textbf{85.49} & 90.86 & 93.54\\
\hline
\end{tabular}
}
\end{center}
\end{table}

As indicated in TABLE \ref{tab:clcd_res}, CFNet achieves an F1 score of 81.41\%, an IoU of 68.65\%, and a recall of 81.08\% on the CLCD dataset, surpassing all classical change detection algorithms. Although the precision is lower than GaMPF’s 84.6\%, CFNet effectively achieves the trade-off between recall and precision, exceeding GaMPF by 12.72\% in recall. Notably, CFNet demonstrates significant improvements in IoU and F1 scores compared to other leading change detection algorithms, with an F1 score increase of 2.52\% and an IoU increase of 3.51\% over EfficientCD. Furthermore, as shown in Table \ref{tab:levir}, CFNet achieves an F1 score of 92.18\% and an IoU of 85.49\% on the LEVIR-CD dataset, with a slight improvement of 0.05\% in F1 and 0.09\% in IoU compared to CGNet. As depicted in Table \ref{tab:sysu}, on the SYSU-CD dataset, CFNet achieves an F1 score of 82.87\%, an IoU of 70.77\%, and a recall of 84.53\%, marking increases of 1.07\% in F1, 2.59\% in IoU, and 4.8\% in recall compared to SSANet. The outstanding performance of CFNet across these three datasets clearly illustrates its effectiveness and robustness in remote sensing change detection task.

\begin{figure*}[!t]
\centering
\includegraphics[width=\textwidth]{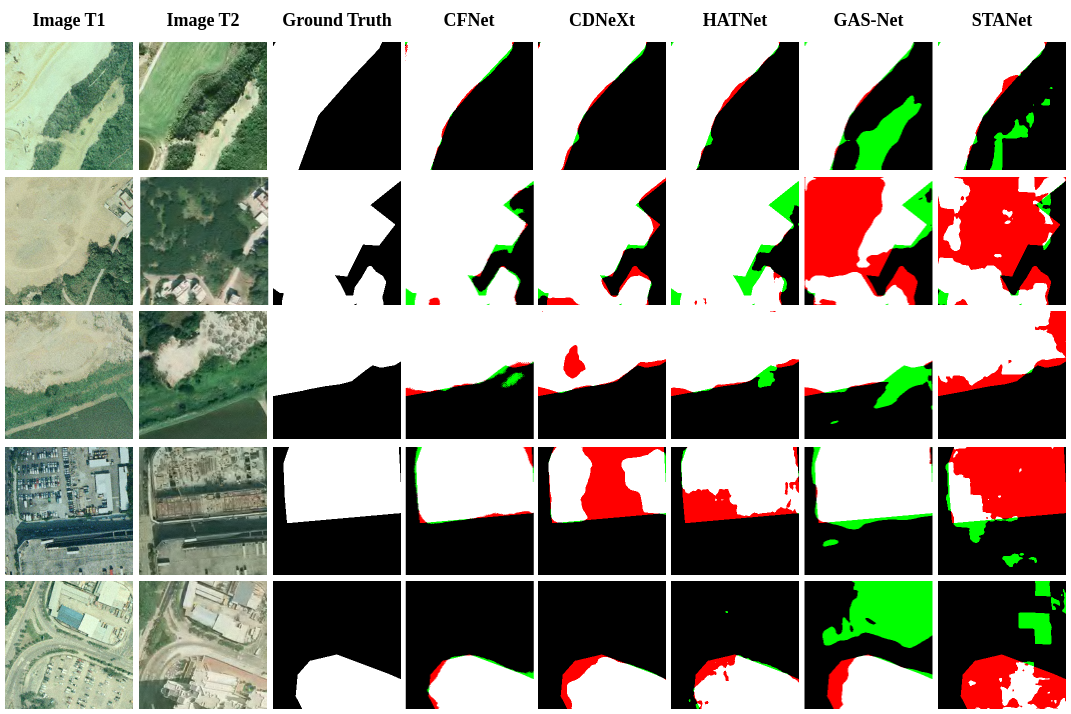}
\caption{The visualization result in SYSU-CD dataset.}
\label{fig:sysu_vis}
\end{figure*}
In Fig.\ref{fig:cl_vis} to \ref{fig:sysu_vis}, the visualization results of CFNet on the CLCD, LEVIR-CD, and SYSU-CD datasets are presented. For each dataset, the visualization includes Image T1, Image T2, Ground Truth, and the predictions made by CFNet, along with the predictions from several top-performing algorithms for comparative analysis. From Figures 5 to 8, it is evident that CFNet demonstrates superior performance across all three datasets compared to other change detection algorithms. Additionally, in the visualization results, we denote true positives (TP) in white, true negatives (TN) in black, false negatives (FN) in red, and false positives (FP) in green.

\begin{table}
\begin{center}
\caption{QUANTITATIVE RESULTS ON SYSU-CD}
\label{tab:sysu}
\setlength{\arrayrulewidth}{0.4mm}
\renewcommand{\arraystretch}{1.3} 
\resizebox{3.3in}{!}{
\begin{tabular}{ c  c  c  c  c }
\hline
Model & F1 & IoU & Rec & Prec\\
\hline
STANet & 72.79 &57.22&66.71&80.08\\
MSCANet &77.33 &63.04 &72.40 &82.99 \\
GAS-Net&78.50&64.61&78.22&78.77\\
BIT &78.65 &64.81 &73.68 &84.34 \\
SNUNet &79.54 &66.02 &75.87 &83.58 \\
P2V &79.73 &66.29 &79.29 &80.17 \\
HCGMNet &79.76 &66.33 &74.15 &86.28 \\
CGNet &79.92 &66.55 &74.37 &86.37\\
HATNet&80.24&67.00&78.23&82.36\\
AFCF3D-Net &80.75 &67.71& 74.16&88.63\\
DARNet &81.03 &68.10 &79.11 &83.04 \\
SSANet &81.08 &68.18 &79.73 &82.48 \\
CDNeXt&81.36 & 68.57& 74.10 & \textbf{90.20}\\
\textbf{CFNet}& \textbf{82.87} & \textbf{70.77} & \textbf{84.53} & 81.31\\
\hline
\end{tabular}
}
\end{center}
\end{table}

\subsection{Ablation Study}

To further validate the effectiveness of the proposed Focuser module and Content-Aware strategy in remote sensing change detection task, we conducted ablation experiments for both Content-Aware and Focuser. The results for IoU and F1 scores on the CLCD, SYSU-CD, and LEVIR-CD datasets are presented in TABLE \ref{tab:iou ablation} and TABLE \ref{tab:f1 ablation}, respectively. In the model where both Content-Aware and Focuser are ablated, the outputs from the two Content Decoder modules are directly fed into the Change Decoder module, and $L_{main}$ serves as the final loss function. In the model with the Focuser module ablated, we retained the output of Focuser module for ``Change Content Collection" and ``Unchanged Content Collection" to ensure the integrity of Content-Aware; however, the outputs $RM_i,i=1,2,3,4$ from the Focuser module were removed, preventing the model from focusing progressively on changed areas within the Change Decoder module. In the model ablated for Content-Aware, we eliminated the processes for computing ``Change Content Collection" and ``Unchanged Content Collection", while using $L_{main}$ as the final loss function. The results from the ablation experiments indicate that both Focuser and Content-Aware effectively enhance the model's performance in change detection task. Moreover, they exhibit a synergistic effect; when both modules are present, the model's performance is further enhanced.
\begin{table*}
\begin{center}
\caption{ABLATION STUDY IN IOU INDEX}
\label{tab:iou ablation}
\setlength{\arrayrulewidth}{0.4mm}
\renewcommand{\arraystretch}{1.3} 
\resizebox{6 in}{!}{
\begin{tabular}{ c  c  c  c  c  c  c }
\hline
Model & Content-Aware & Focuser & CLCD & SYSU-CD & LEVIR-CD \\
\hline
\multirow{4}*{CFNet} & \usym{2613} &\usym{2613} & 62.76 & 68.25&84.19 \\
~ & \usym{2713} &\usym{2613} & 63.85  & 68.36& 84.54\\
~ & \usym{2613} &\usym{2713} & 64.65 & 68.94& 84.90\\
~ & \usym{2713} &\usym{2713} & \textbf{68.65} & \textbf{70.77}& \textbf{85.49}\\
\hline
\end{tabular}
}
\end{center}
\end{table*}

\begin{table*}
\begin{center}
\caption{ABLATION STUDY IN F1 INDEX}
\label{tab:f1 ablation}
\setlength{\arrayrulewidth}{0.4mm}
\renewcommand{\arraystretch}{1.3} 
\resizebox{6 in}{!}{
\begin{tabular}{ c  c  c  c  c  c  c }
\hline
Model & Content-Aware & Focuser & CLCD & SYSU-CD & LEVIR-CD \\
\hline
\multirow{4}*{CFNet} & \usym{2613} &\usym{2613} & 77.12 & 81.13&91.42 \\
~ & \usym{2713} &\usym{2613} & 77.94  & 81.21& 91.62\\
~ & \usym{2613} &\usym{2713} & 78.53 & 81.62& 91.83\\
~ & \usym{2713} &\usym{2713} & \textbf{81.41} & \textbf{82.89}& \textbf{92.18}\\
\hline
\end{tabular}
}
\end{center}
\end{table*}

To analyze the impact of the loss weight ratio on model performance, we conducted another ablation study by varying the ratio of $\alpha$ to $\beta$ (where $\beta$ and $\gamma$ are set to the same value to maintain a consistent ratio between $L_{cc}$ and $L_{ucc}$). Table~\ref{tab:ablation} presents the results of this study. Our original model setting corresponds to a ratio of $10:1$. In our original model, we set $\alpha=1$ and $\beta=\gamma=0.1$, corresponding to the $10:1$ ratio in the table. This choice was based on observing the absolute magnitudes of $L_{main}$, $L_{cc}$, and $L_{ucc}$, ensuring they remain on a similar scale to prevent any single component from excessively dominating the model’s fitting process. From Table~\ref{tab:ablation}, we observe that increasing the ratio of $\alpha$ results in $L_{main}$ dominating the model, thereby reducing the influence of $L_{cc}$ and $L_{ucc}$. This leads to a decline in model performance. Conversely, when increasing the ratio of $\beta$ and $\gamma$, the auxiliary losses ($L_{cc}$ and $L_{ucc}$) overly influence the fitting process, preventing the model from effectively learning the ground truth information through $L_{main}$. This causes the model's performance to degrade even more rapidly. The results indicate that a well-balanced loss ratio is crucial for achieving optimal performance.

\begin{table}[h]
    \begin{center}
    \caption{Ablation study on the $\alpha$:$\beta$($\gamma$) ratio}
    \label{tab:ablation}
    \setlength{\arrayrulewidth}{0.4mm}
    \renewcommand{\arraystretch}{1.3} 
    \resizebox{3.2 in}{!}{
    \begin{tabular}{c c c c c}
        \hline
        $\alpha$:$\beta$($\gamma$) & IoU & F1 & Rec & Prec \\
        \hline
        100:1 & 62.64 & 77.78 & 76.87 & 78.72\\
        50:1  & 63.84 & 78.00 & 79.62 & 76.31 \\
        \textbf{10:1}  & \textbf{68.65} & \textbf{81.41} & \textbf{81.08} & \textbf{81.74} \\
        1:1   & 60.88 & 75.68 & 74.71 & 76.69 \\
        1:5   & 58.92 & 74.15 & 74.00 & 74.29 \\
        1:10  & 56.24 & 71.99 & 71.64 & 72.35 \\
        \hline
    \end{tabular}
    }
    \end{center}
\end{table}

\section{Discussion}
\subsection{Focuser Module in CFNet}

\begin{figure*}[!t]
\centering
\includegraphics[width=\textwidth]{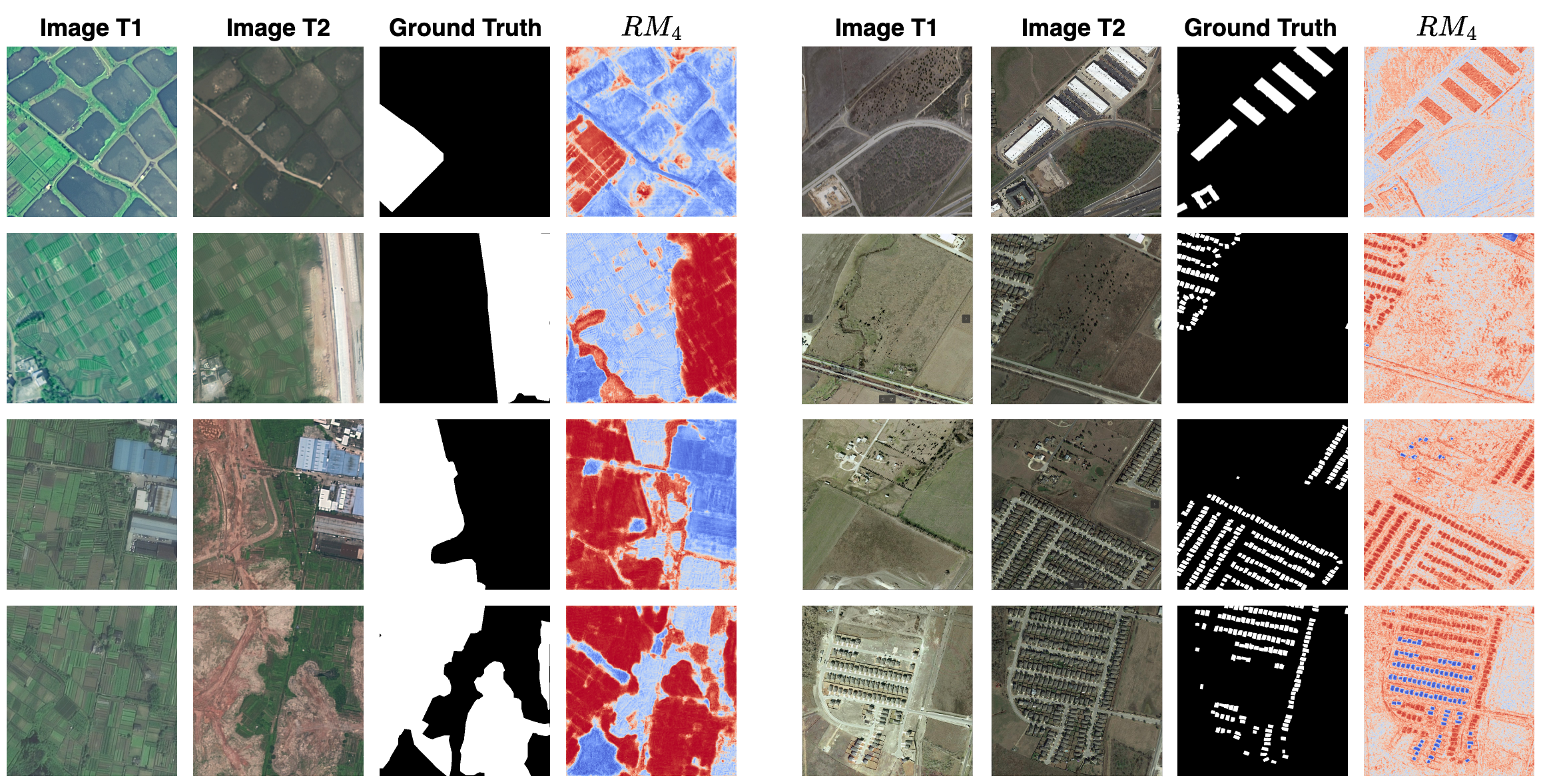}
\caption{The visualization of $RM_4$ from the test dataset of CLCD and LEVIR-CD. The left half demonstrates results on CLCD and the right half demonstrates results on LEVIR-CD. In the visualization of $RM_4$, pixels with values closer to 1 are represented in red, indicating a higher likelihood of being a changed area, while pixels with values closer to 0 are represented in blue, indicating a higher likelihood of being an unchanged area.}
\label{fig:f_disc}
\end{figure*}

In this study, we employed the Focuser module to separately generate the ``Change Content Collection'' and the ``Unchanged Content Collection.'' The Focuser module plays a critical role in our framework by distinguishing between areas that exhibit changes and those that remain constant, which is essential for the effectiveness of the change detection task. A pivotal step in the Focuser module involves the computation of $RM_i$, where $i=1,2,3,4$. Among these, $RM_4$ is particularly noteworthy due to its proximity to the final Change Map within the network architecture. As the last stage in this process, $RM_4$ holds the most significant influence on the model's performance in identifying changed areas accurately.

To evaluate the effectiveness of the Focuser module in enhancing the change detection capability, we conducted an extensive visualization analysis of $RM_4$. This visualization is critical as it provides direct insights into how well the Focuser module aligns with the $\text{Ground Truth}$ and contributes to the overall learning process of the network. Specifically, we selected $RM_4$ for analysis because it encapsulates the refined features and predictions generated by preceding modules, thus serving as the most representative output for assessing the performance of the Focuser module. The results of this visualization are depicted in Fig.~\ref{fig:f_disc}, where we present the outputs of $RM_4$ derived from the test dataset of CLCD and LEVIR-CD. In these visualizations, pixels are assigned values within the range of $[0,1]$, where higher values, closer to 1, are rendered in red to signify a greater likelihood of corresponding to changed areas. Conversely, pixels with lower values, closer to 0, are shown in blue, indicating a higher probability of being part of unchanged areas. This intuitive color mapping enables a clear interpretation of the model's predictions.

The visualization results of $RM_4$ demonstrate a remarkable alignment with the $\text{Ground Truth}$, highlighting the efficacy of the Focuser module in accurately distinguishing between changed and unchanged areas. The ability of $RM_4$ to closely approximate the $\text{Ground Truth}$ underscores the module's capacity to extract and refine discriminative features relevant to the change detection task. Furthermore, the clear separation between changed and unchanged areas in the visualizations validates the module's robustness in enhancing the overall network's ability to fit the $\text{Ground Truth}$, thereby improving the precision of the change detection outcomes.

\subsection{Content-Aware Strategy in CFNet}
Traditional change detection algorithms often overlook the impact of style differences between bi-temporal images on model performance, especially given the strong fitting capabilities of DNN. In this study, inspired by self-similarity and the human ability to recognize content through internal structural features, we propose the Content-Aware strategy. In addition to the main branch where the model fits the ground truth, we introduce two auxiliary branches that decode content features. These branches leverage the previously mentioned $L_{cc}$ and $L_{ucc}$ to guide the model in learning content features for both changed and unchanged areas, thereby imposing stronger constraints on the main branch’s fitting of the changed areas. 

\begin{figure*}[!t]
\centering
\includegraphics[width=\textwidth]{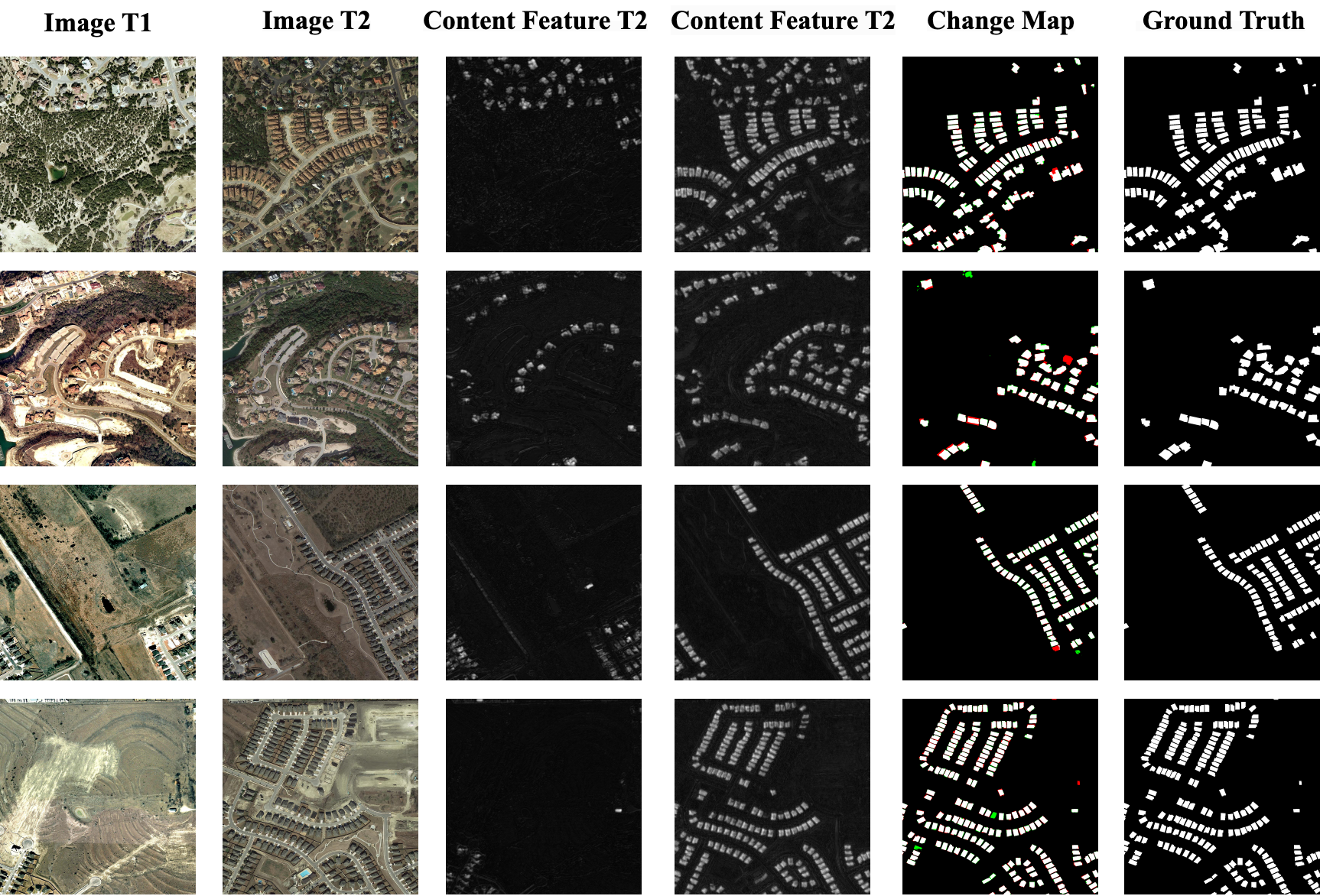}
\caption{Visualization of the Content Decoder’s output on the LEVIR-CD dataset. Each row presents a bi-temporal image pair (first and second columns), the corresponding grayscale feature maps outputed by the Content Decoder (third and fourth columns), the final predicted ``Change Map" and the ground truth. Notably, since the Content Decoder is also influenced by $L_{main}$ and the LEVIR-CD dataset primarily focuses on building changes, the content features predominantly capture structural information related to buildings.}
\label{fig:content_res}
\end{figure*}

To further illustrate the effectiveness of the Content-Aware strategy in extracting content features, we visualize the largest-scale feature maps output by the Content Decoder. As direct visualization of all feature channels is not feasible, we compute the channel-wise mean and normalize the values to the [0,255] range to generate grayscale images. Despite the inevitable loss of some feature details, the resulting images provide an intuitive understanding of how the Content Decoder captures content information while filtering out style differences.

Fig.\ref{fig:content_res} presents four randomly selected cases from the LEVIR-CD dataset. In each row, the first and second columns are the input bi-temporal images, which exhibit noticeable style variations. The third and fourth columns show the corresponding grayscale representations of the largest-scale feature maps output by the Content Decoder. These feature maps effectively suppress style differences while preserving structural content, highlighting the Content Decoder’s ability to output content features.(It is important to note that the Content Decoder is also influenced by $L_{main}$. Since the LEVIR-CD dataset primarily focuses on building changes, the content features predominantly represent structural information related to buildings.) This ensures that the subsequent Change Decoder operates on more consistent content representations, thereby improving the accuracy of change detection. 

However, in this study, the ratio of $\alpha$ and $\beta$, which defines the overall loss, was manually determined by observing the behavior of $L_{main}$, $L_{cc}$, and $L_{ucc}$ during training. Therefore, it remains an open question how to set this ratio more scientifically, or even enable the model to flexibly adjust the ratio at different stages of training. I believe that a more rigorous and adaptable adjustment of $\alpha$ and $\beta$ will further enhance the performance of CFNet in remote sensing image change detection task.


\section{Conclusion}
In this paper, we presented CFNet, a novel framework for remote sensing change detection that addresses the challenges posed by style variations between bi-temporal images. By focusing on content features, CFNet reduces the impact of unpredictable style differences, which often hinder the accuracy of change detection models. The introduction of the Content-Aware strategy enhances the model's ability to capture intrinsic content features, while the Focuser module allows dynamic emphasis on both changed and unchanged areas throughout the detection process.

Our extensive experiments on CLCD, LEVIR-CD, and SYSU-CD demonstrate that CFNet consistently outperforms existing state-of-the-art methods. Notably, CFNet achieved significant improvements in F1 score and Intersection over Union (IoU), illustrating its robustness and generalizability across diverse remote sensing scenarios. Furthermore, our ablation studies highlight the complementary roles of the Content-Aware strategy and the Focuser module in enhancing detection accuracy.

Future work could explore more adaptive mechanisms for balancing the loss components of CFNet, potentially enabling further improvements in model performance. Moreover, expanding the model’s applicability to other remote sensing domains, such as environmental monitoring and disaster response, represents a promising direction for future research.

\section*{Acknowledgment}
The numerical calculations in this article have been done
on the supercomputing system at the Supercomputing Center,
Wuhan University, Wuhan, China.


\bibliographystyle{IEEEtran}
\bibliography{IEEEabrv,ref}

\newpage
\section{Biography Section}

\begin{IEEEbiography}[{\includegraphics[width=1in,height=1.25in,clip,keepaspectratio]{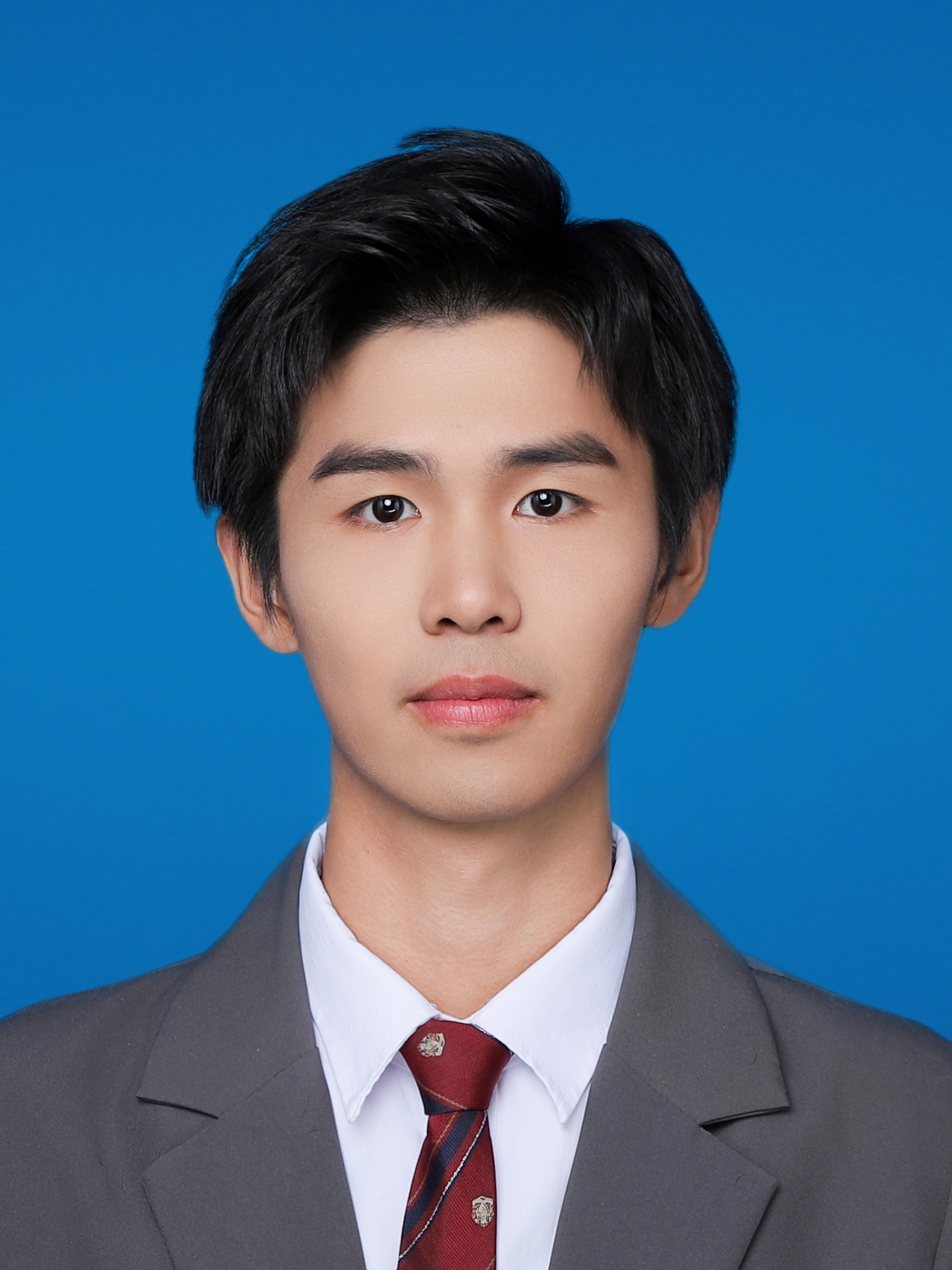}}]{Fan Wu}
is currently pursuing the bachelor's degree with Remote Sensing Information Engineering Institute, Wuhan University, Wuhan, China. His research interests include computer vision and remote sensing image processing.
\end{IEEEbiography}

\begin{IEEEbiography}[{\includegraphics[width=1in,height=1.25in,clip,keepaspectratio]{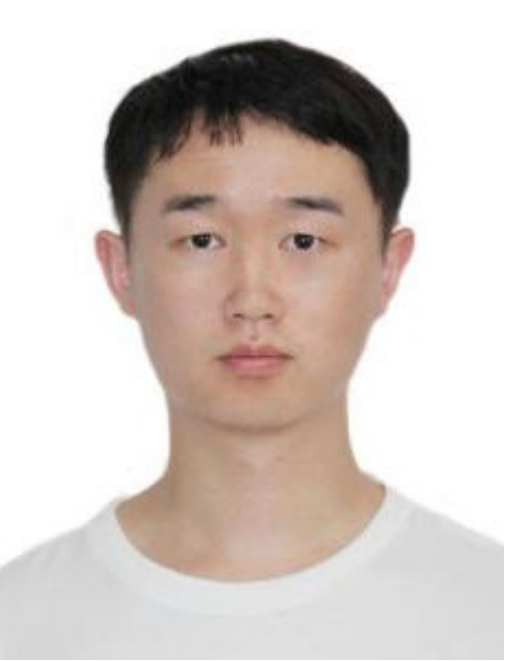}}]{Sijun Dong}
received the bachelor’s degree in
computer science and technology from Guangxi
University, Nanning, China, in 2017, and the master’s degree in computer science and technology
from Shenzhen University, Shenzhen, China, in
2020. He is currently pursuing the Ph.D. degree in
remote sensing science and technology with Wuhan
University, Wuhan, China, with a research focus on
computer vision and remote sensing image processing, under the supervision of Prof. Xiaoliang Meng.
\end{IEEEbiography}

\begin{IEEEbiography}[{\includegraphics[width=1in,height=1.25in,clip,keepaspectratio]{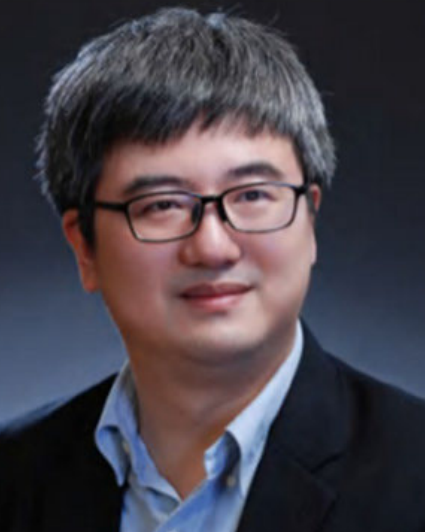}}]{Xiaoliang Meng} received the Ph.D. degree from Wuhan University, Wuhan, China, in 2009. He was
a Visiting Scholar and a Post-Doctoral Scientist in USA for three years and participated in the NASA
ICCaRS Project. He is currently a Distinguished Professor at School of Remote Sensing and Information Engineering, Wuhan University. His main research interest is intelligent geospatial sensing. Dr.Meng has received the “Best Young Authors Award” from the International Society
for Photogrammetry and Remote Sensing (ISPRS).
\end{IEEEbiography}

\end{document}